\pdfoutput=1

\documentclass[11pt]{article}
\PassOptionsToPackage{table}{xcolor}
\usepackage{latexml}
\iflatexml
  \usepackage[a4paper,margin=2.5cm]{geometry}
  \usepackage{natbib}
\else
  \usepackage[final]{acl}
\fi
\usepackage{times}
\usepackage{latexsym}
\usepackage{microtype}
\usepackage{inconsolata}
\usepackage{graphicx}
\usepackage{booktabs}
\usepackage{amsmath,amssymb}
\usepackage{multirow}
\usepackage{xspace}
\usepackage{array}
\usepackage{float}
\usepackage{placeins}
\usepackage[skins,breakable]{tcolorbox}
\usepackage{url}

\definecolor{rowhi}{RGB}{232,240,251}    
\definecolor{cellhi}{RGB}{229,243,233}   
\definecolor{cellpos}{RGB}{217,237,222}  
\definecolor{cellneg}{RGB}{249,229,229}  
\definecolor{cellneu}{RGB}{240,240,240}  
\definecolor{promptbg}{RGB}{246,246,246} 
\definecolor{promptrule}{RGB}{210,210,210}

\newtcolorbox{promptbox}[1][]{%
  enhanced, breakable, sharp corners,
  colback=promptbg, colframe=promptrule,
  boxrule=0.4pt, left=5pt, right=5pt, top=4pt, bottom=4pt,
  fontupper=\small\ttfamily, #1
}

\newcommand{\rg}{\textsc{Read-Gate}\xspace}
\newcommand{\pdisc}{\ensuremath{P_{\mathrm{disc}}}\xspace}
\newcommand{\ppost}{\ensuremath{P_{\mathrm{post}}}\xspace}
\newcommand{\edisc}{\ensuremath{\mathcal{E}_{\mathrm{disc}}}\xspace}
\newcommand{\epost}{\ensuremath{\mathcal{E}_{\mathrm{post}}}\xspace}
\newcommand{\eretr}{\ensuremath{\mathcal{E}_{\mathrm{retr}}}\xspace}
\newcommand{\eamb}{\ensuremath{\mathcal{E}_{\mathrm{amb}}}\xspace}
\newcommand{\ewrong}{\ensuremath{\mathcal{E}_{\mathrm{wrong}}}\xspace}

\newcommand{\maybegraphic}[2][]{%
  \IfFileExists{#2}{\includegraphics[#1]{#2}}{%
    \fbox{\parbox{0.86\linewidth}{\centering Missing figure file: \texttt{\detokenize{#2}}}}%
  }%
}

\iflatexml
  \title{Before Reasoning Can Fail: Pre-Evidence Procedural Failures in Agentic RAG}
  \author{Daeyoung Roh\\
    Independent Researcher\\
    \texttt{dybroh@gmail.com}
    \and
    Donghee Han\\
    KAIST\\
    \texttt{handonghee@kaist.ac.kr}}
  \date{}
\else
  \title{Before Reasoning Can Fail: Pre-Evidence Procedural Failures in Agentic RAG\thanks{Code is available at \url{https://github.com/Noverse0/before-reasoning-fails}}}
  \author{
    Daeyoung Roh \\
    Independent Researcher \\
    \texttt{dybroh@gmail.com} \\\And
    Donghee Han \\
    KAIST \\
    \texttt{handonghee@kaist.ac.kr} \\
  }
\fi

\begin{document}
\maketitle
\iflatexml
  \begin{center}
    \small Code: \url{https://github.com/Noverse0/before-reasoning-fails}
  \end{center}
\fi

\begin{abstract}
Agentic retrieval-augmented generation (RAG) systems can fail before
evidence-conditioned reasoning is tested: an agent may retrieve candidate
snippets but finalize without inspecting them. We study this failure mode as
a procedural property of the agent trajectory, decomposing wrong answers
into pre-evidence discipline failures and post-gold-read failures using
saved tool-call traces, retrieved evidence, read passages, and final
answers. Across 12{,}000 paired trajectories on HotpotQA,
2WikiMultiHopQA, and MuSiQue, the two failure types are largely
non-redundant: the both-trigger rate is in $[11.2\%, 13.1\%]$ across regex and spaCy entity extractors. We then
evaluate \rg, a minimal runtime invariant requiring an agent to read after
search and before finalization. Forced reading improves LLM-Acc by
14.9--19.9 points on trajectories that would otherwise skip reading and by
3.2--9.4 points on full minimal-reasoning cells. Additional diagnostics
show that larger hidden thinking budgets do not necessarily increase
evidence inspection. Together, these results indicate that
evidence-gathering should be evaluated as a trajectory-level control
problem, separately from answer-side reasoning.\end{abstract}

\section{Introduction}

\begin{figure}[t]
    \centering
    \maybegraphic[width=\linewidth]{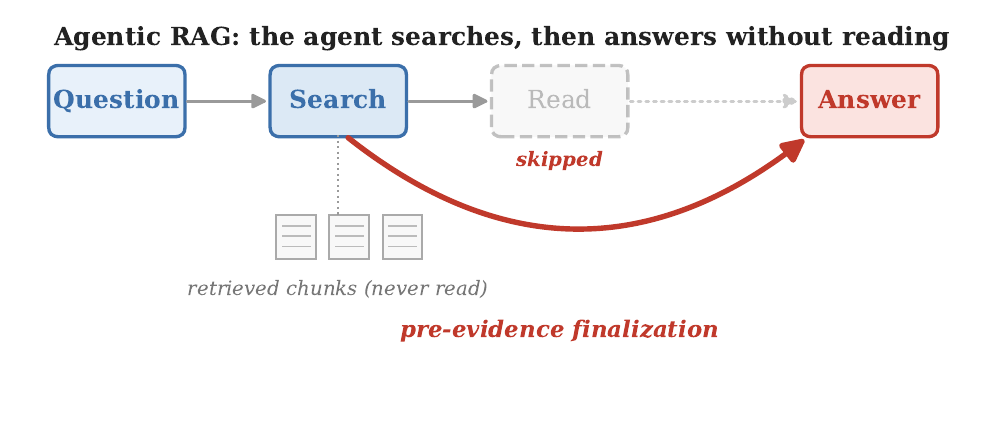}
    \caption{Pre-evidence finalization in agentic RAG. The agent retrieves
    candidate chunks with snippets but finalizes without executing a read
    action, so the final answer is produced before evidence inspection.}
    \label{fig:intro-trace}
\end{figure}

Agentic retrieval-augmented generation (RAG) systems can fail before
evidence-conditioned reasoning ever begins. In multi-hop question answering,
we observe agents that issue a search, receive plausible retrieved snippets,
and finalize an answer without reading any retrieved passage
(Figure~\ref{fig:intro-trace}). Such trajectories are not ordinary reasoning
failures: the model has not yet inspected the evidence on which its answer is
supposed to depend.

This failure mode is specific to agentic RAG. Unlike fixed-context RAG
systems, agentic systems decide when to search, which retrieved results to
inspect, which passages to read, and when to stop and answer
\citep{nakano2021webgpt,yao2023react,schick2023toolformer,qin2023toolllm}.
This flexibility is useful for multi-hop questions, where the required
evidence is often distributed across passages
\citep{yang2018hotpotqa,ho2020constructing,trivedi2022musique}, but it also
introduces a procedural risk. An agent may retrieve relevant evidence yet
answer from a snippet, read only part of the evidence chain, or terminate
without reading at all. Final-answer accuracy alone does not reveal whether
the agent actually executed the evidence-gathering procedure before answering
\citep{liu2023agentbench}.

We argue that these failures should be separated from post-gold-read
failures. If a model reads the necessary evidence and still answers
incorrectly, the system may need stronger reasoning, better synthesis, or
better supervision. If the model never reads the evidence, the problem is
different: the execution procedure allowed the agent to stop too early.
Treating both cases as generic answer errors hides an actionable source of
failure in agentic RAG.

To make this distinction measurable, we introduce a trajectory-level
decomposition of wrong answers into failures of evidence inspection and
failures after evidence has been read. The decomposition is operational: it is
computed from tool calls, read counts, retrieved snippets, read passages, gold
evidence annotations when available, and final-answer correctness. We call
wrong trajectories with insufficient evidence inspection \emph{discipline
failures}, where ``discipline'' refers to procedural compliance with the
evidence-inspection protocol rather than answer-side reasoning. We call
wrong trajectories after reading at least some gold-supporting evidence
\emph{post-gold-read failures}. These labels are not claims about latent model cognition; they
are deterministic measurements over saved agent traces.

This framing suggests a simple intervention. If some errors occur because the
agent answers before reading, then increasing reasoning effort is not the only
relevant fix. A runtime constraint may recover failures that a stronger or
more deliberative model would not necessarily prevent. We therefore evaluate
\rg, a minimal environment-level invariant: after search, the agent must read
before it can answer. If the model attempts to finalize without reading, the
environment rejects the action and returns a corrective observation
prompting the agent to read a retrieved chunk. The intervention does not change the model, retriever, index, judge, or
reasoning budget.

We evaluate this hypothesis on HotpotQA, 2WikiMultiHopQA, and MuSiQue using
12{,}000 paired OpenAI-controller trajectories. The trajectory decomposition
shows that discipline failures and post-gold-read failures are largely
non-redundant: under multi-label reclassification, the both-trigger
rate is in $[11.2\%, 13.1\%]$ across regex and spaCy entity extractors. \rg directly targets the pre-evidence regime. On
examples with zero-read finalization under the voluntary minimal-reasoning
policy, forced reading improves LLM-Acc by 14.9--19.9 points; on full
minimal-reasoning cells, \rg improves LLM-Acc by 3.2--9.4 points. Additional
diagnostics show that evidence inspection is not simply a byproduct of
larger hidden thinking budgets. A context-injection control further shows
that delivering the same chunk text without a read action recovers only a
minority of \rg's gain.

Our contributions are threefold:
\begin{itemize}
    \item We operationalize pre-evidence procedural failures in agentic RAG
    through a trajectory-level decomposition that separates discipline
    failures from post-gold-read failures.

    \item We introduce \rg, a minimal runtime invariant that prevents
    finalization before reading retrieved evidence, without changing the
    model, retriever, judge, or reasoning budget.

    \item Across three multi-hop QA benchmarks, we show that the two failure
    types are largely non-redundant, that \rg recovers zero-read failures
    when residual discipline error is high, and that evidence inspection is
    a distinct control axis from answer-side reasoning, not guaranteed by
    increased reasoning effort alone.
\end{itemize}
\section{Related Work}

\paragraph{Agentic RAG and retrieval control.}
RAG combines parametric language models with non-parametric evidence retrieval
for knowledge-intensive generation \citep{lewis2020retrieval}. Agentic systems
such as WebGPT, ReAct, Toolformer, and ToolLLM interleave model reasoning with
external tool actions
\citep{nakano2021webgpt,yao2023react,schick2023toolformer,qin2023toolllm};
retrieval-oriented systems such as IRCoT, FLARE, and A-RAG study how agents
search, retrieve, and read evidence during multi-step QA
\citep{trivedi2023ircot,jiang2023active,du2026arag}. These works make retrieval
and tool use controllable, but they do not isolate the execution-time case
where retrieval is invoked and then bypassed before evidence inspection.

\paragraph{Policy learning versus runtime constraints.}
Self-RAG, CRAG, and recent RL-based search agents train or adapt retrieval
policies through reflection tokens, learned evaluators, or reinforcement
learning
\citep{asai2024selfrag,yan2024crag,jin2025searchr1,chen2025research,song2025r1searcher,wang2025erase}.
\rg instead imposes a runtime invariant on the environment: after retrieval,
finalization can be rejected until a read action occurs. Related calibration
and retrieval-triggering methods estimate when model knowledge is insufficient
from confidence, uncertainty, or answer-level signals
\citep{kadavath2022know,lin2022teaching,mallen2023whennot,press2023selfask,kuhn2023semantic,yoran2024making}.
Our trigger is different: it is an observable trajectory violation in which the
agent has retrieved candidates but has not inspected them before answering.

\paragraph{Evaluation, grounding, and reasoning effort.}
Attribution, faithfulness, RAG evaluation, and LLM-as-judge frameworks assess
whether final answers are correct or supported by evidence
\citep{rashkin2023measuring,gao2023alce,liu2023verifiability,es2024ragas,saadfalcon2024ares,zheng2023judging,liu2023geval},
while agent benchmarks show that final outcomes can mask tool-use failures
\citep{liu2023agentbench}. Reasoning methods such as chain-of-thought,
self-consistency, process supervision, and verification improve or supervise
answer-side reasoning
\citep{wei2022chain,wang2023selfconsistency,kojima2022large,lightman2023lets,dhuliawala2024chain}.
We instead evaluate the external action sequence itself: whether the agent
executes evidence inspection before producing the final answer.
\section{Framework: Pre-Evidence Failures in Agentic RAG}
\label{sec:framework}

We now make the preceding distinction operational. An agent trajectory consists
of a sequence of tool actions, including \texttt{search}, \texttt{read}, and
\texttt{final}, together with retrieved snippets, read passages, gold evidence
annotations when available, and final-answer correctness. Our goal is not to
infer latent model intent, but to label observable failure modes in the saved
trajectory. The framework applies to agents that expose such discrete
\texttt{search}/\texttt{read}/\texttt{final} tool actions; agents that
interleave retrieval and generation implicitly, with no observable action
boundary, are outside its scope (see Limitations).

\begin{figure}[t]
    \centering
    \maybegraphic[width=\linewidth]{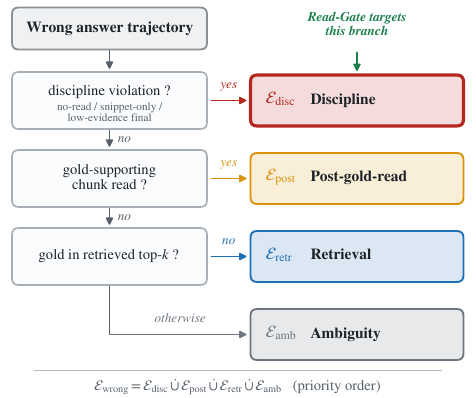}
    \caption{Trajectory-level decomposition of wrong answers. Each wrong
    trajectory is assigned to a discipline, post-gold-read, retrieval, or
    residual ambiguity branch. \rg targets the discipline branch by blocking
    finalization before evidence inspection.}
    \label{fig:trajectory-decomposition}
\end{figure}

We define two primary error axes for agentic RAG
(Figure~\ref{fig:trajectory-decomposition}). The first is
\emph{discipline}: whether the agent follows the evidence-inspection
procedure before answering. The second is \emph{post-gold-read error}:
whether the agent still answers incorrectly after reading at least some
gold-supporting evidence. This distinction separates failures that occur
before evidence-conditioned reasoning can be tested from failures that remain
after evidence has been inspected.

\subsection{Trajectory-Level Error Decomposition}
\label{sec:trajectory-decomposition}

For mutually exclusive accounting, we assign each wrong trajectory to the
first matching bucket under a fixed priority order:
\[
\ewrong
=
\edisc \;\dot{\cup}\; \epost \;\dot{\cup}\; \eretr \;\dot{\cup}\; \eamb .
\]
The disjointness is therefore operational rather than semantic: a trajectory
may satisfy multiple failure indicators, but the priority rule assigns it to
exactly one accounting bucket. Discipline checks fire first, followed by
post-gold-read, retrieval, and residual ambiguity. Section~\ref{sec:framework-multilabel}
separately analyzes independent multi-label indicators.

\subsection{Operational Failure Indicators}
\label{sec:failure-indicators}

\paragraph{Discipline failure.}
\edisc fires when the answer is wrong and the trajectory violates a procedural
evidence-inspection condition. We use ``discipline'' in the procedural sense:
whether the agent follows the required evidence-inspection protocol before
answering. We use three observable subtypes:
\begin{itemize}
    \item \textbf{No-read final}: the agent emits a final answer with
    \texttt{read\_count}=0.
    \item \textbf{Snippet-only final}: an answer entity appears only in
    retrieved search snippets, never in any read chunk.
    \item \textbf{Low-evidence final}: named entities from the question are
    covered in read chunks at less than an 80\% rate.
\end{itemize}
Entity coverage is computed with a deterministic named-entity matching
heuristic; implementation details are given in Appendix~\ref{app:entity-labeling}.
Because snippet-only and low-evidence subtypes rely on surface matching
heuristics, we also report a strict no-read-only lower bound and threshold
sweeps in the robustness analyses. Pooled over the main cells, no-read final
is the majority subtype and the only one that requires no entity-matching
heuristic at all; we therefore treat it as the primary discipline signal and
the other two as secondary diagnostics (Appendix~\ref{app:subtype-decomposition}).

\paragraph{Post-gold-read failure.}
\epost fires when retrieval succeeded, the agent read at least one
gold-supporting chunk, and the final answer is still wrong. This category is
narrower than complete evidence-conditioned answering: in multi-hop QA,
reading one gold chunk may still leave the supporting chain incomplete.
Because this category depends on gold supporting evidence annotations,
datasets without complete gold evidence fields support only limited \epost
analysis (see \S\ref{sec:datasets} for the dataset-specific caveat).

\paragraph{Retrieval and ambiguity.}
\eretr captures cases where gold evidence does not appear in any top-$k$
retrieval result. \eamb is the residual bucket for cases that do not cleanly
fit the operational definitions. In robustness analysis, residual ambiguity
remains below 15\% of wrong cases.

\subsection{\rg: A Runtime Evidence-Inspection Constraint}
\label{sec:readgate}

The decomposition suggests a direct diagnostic intervention. If a wrong
trajectory occurs because the agent finalizes before inspecting evidence, then
the environment can enforce a minimal evidence-inspection invariant: after
search, the agent must read before it can answer.

\rg is a deterministic environment rule that enforces a single
\emph{read-before-final} invariant: if the model emits \texttt{final} with
\texttt{read\_count}=0, the environment rejects the action and returns a
corrective observation instructing the agent to call \texttt{read} on a
promising chunk before answering.

\rg operates at inference time. It does not change model weights,
retrieval, the index, the judge, decoding temperature, or reasoning budget.
A stricter checkpoint that additionally rejects a non-\texttt{read} action
issued immediately after \texttt{search} (\emph{read-after-search}) is
implemented within the same corrective-hint framework but is not part of the
main \rg condition used throughout the paper; Appendix~\ref{app:gate-family}
explores a family of stronger procedural gates in this spirit. Operational
details for the main \rg invariant (trigger predicate, corrective hint,
rollout protocol) are in Appendix~\ref{app:rg-operational}.

\subsection{Priority Accounting and Multi-Label Indicators}
\label{sec:framework-multilabel}

Priority assignment is useful for mutually exclusive accounting, but it can
mask co-occurring labels. We therefore also compute independent binary
indicators for discipline and post-gold-read failure. We write \pdisc{} and
\ppost{} for the corresponding trajectory-level event probabilities,
estimated as cell-level rates in aggregate analyses and modeled with
question-clustered logistic regressions in Section~\ref{sec:exp-differential}.
This multi-label version is the basis for the both-trigger overlap test in
Section~\ref{sec:exp-validity}. Robustness checks further report a strict
no-read-only lower bound and vary the low-evidence threshold from 0.6 to 0.9.

\subsection{Predictions}
\label{sec:predictions}

The framework yields three falsifiable predictions. (H1)
\emph{Non-redundancy}: discipline and post-gold-read labels should not
collapse to a single failure axis. (H2) \emph{Conditional benefit}: \rg's
accuracy gain should scale with residual discipline error rather than with
reasoning level alone. (H3) \emph{Differential response}: \rg{} and reasoning
effort should produce distinguishable signatures over $(\pdisc,\ppost)$,
indicating distinct control axes. We test these predictions through multi-label
reclassification, cell-level correlation, and question-clustered logistic
models with trajectory-level label permutation.

H1's overlap test requires per-chunk gold evidence and is therefore
established on HotpotQA and 2WikiMultiHopQA, the two datasets with such
annotations; MuSiQue, which lacks gold chunk IDs, instead tests H2 alongside
the other two datasets, with its discipline failures corroborated by a
structural, annotation-free signal (90.4\% strict no-read on the most severe
cell; Appendix~\ref{app:subtype-decomposition}).
\section{Experimental Setup}
\label{sec:setup}

\subsection{Datasets}
\label{sec:datasets}

We evaluate on three Wikipedia-style multi-hop QA datasets: HotpotQA
\citep{yang2018hotpotqa}, 2WikiMultiHopQA \citep{ho2020constructing}, and
MuSiQue \citep{trivedi2022musique}. Unless otherwise noted, all main table cells use $n=1{,}000$ examples per dataset--condition pair, paired by question ID across conditions; this includes the OpenAI controllers, the Gemini 2.5 Flash thinking-budget diagnostic, and the prompt-only and ctx-inject mechanism
controls. Boundary ablations, including medium-reasoning \rg checks and
the gate-family probe (Appendix~\ref{app:gate-family}), use matched
$n=100$ samples and are interpreted only as scope checks.

One measurement caveat affects the decomposition. The post-gold-read bucket
requires \emph{per-chunk} gold supporting-evidence annotations, but the
processed MuSiQue export does not contain complete per-chunk gold-evidence
fields. We therefore use MuSiQue for aggregate accuracy, discipline failures,
and \rg effects, but restrict analyses requiring \epost or chunk-level
gold-read quantities to HotpotQA and 2WikiMultiHopQA.

\subsection{Agent Interface and Retrieval Tools}
\label{sec:agent-interfaces}

The main experiments use a two-tool agent interface with hybrid search and
chunk read. The hybrid search tool combines BM25 and dense retrieval using
reciprocal-rank fusion with $k=60$ and returns the top five chunk IDs with
short snippets. The read tool expands a selected chunk ID to the full chunk
text. Cross-search chunk-ID deduplication is enabled, so repeated chunk IDs
are not repeatedly charged as new evidence.

\subsection{Controller Models and Reasoning Levels}
\label{sec:controllers}

The main experiments use OpenAI gpt-4o-mini and gpt-5-mini with minimal and
medium reasoning effort, which separates backbone choice from reasoning-effort
choice within gpt-5-mini. We use OpenAI controllers for the main
within-provider analysis because they provide structured tool-call traces and
multiple reasoning-effort settings for the same model. The 12{,}000 OpenAI
trajectories come from three datasets and four conditions per dataset
(1{,}000 questions each): (i) gpt-4o-mini, (ii) gpt-5-mini minimal, (iii)
gpt-5-mini medium, and (iv) gpt-5-mini minimal plus \rg. Gemini 2.5 Flash is
used only as an external diagnostic for hidden thinking-budget behavior.

\subsection{Evaluation Protocol}
\label{sec:evaluation-protocol}

All main cells use the same retrieval and loop configuration unless otherwise
stated: a 0.6B dense embedding model specified in Appendix~\ref{app:prompts},
hybrid top-$k=5$, maximum loop budget 10, maximum token budget 128k, and
temperature 0.0. The Gemini thinking-budget runs use the same search/read API
and temperature 0.0; the thinking-budget comparison holds the agent loop fixed
and varies only the model's hidden thinking budget.

We report LLM-Acc as the primary outcome using a fixed gpt-5-mini judge at
temperature 0.0; the judge sees only the question, gold answer, and prediction
and returns a binary semantic-equivalence label. Contain-Acc is reported as a
secondary metric where short-form gold answers are available. A cross-family
Gemini 2.5 Pro re-judge of a stratified $n=450$ sample agrees at $\kappa=0.924$
with direction-preserving \rg{} gains on all three datasets
(Appendix~\ref{app:cross-judge}). Full prompts and tool descriptions are in
Appendix~\ref{app:prompts}.

Paired analyses use shared question IDs between conditions. We report exact
McNemar tests for paired correctness changes and use question-clustered
standard errors in logistic models to account for repeated evaluation of the
same question under different conditions.

\paragraph{Reproducibility.}
At the URL in the title footnote, we release the full agent-loop
and \rg implementation, the reproduction scripts used to generate every table
and figure in this paper, and a corpus of 33{,}950 raw agent trajectories
(49 files), including the 12{,}000-trajectory paired corpus with per-question
discipline and post-gold-read labels; the trajectories back every
per-question table except two externally-sourced A-RAG comparison rows,
documented in the release. The release also includes cached analysis outputs
that reproduce every reported number without further API access.
\section{Experiments}
\label{sec:experiments}

We evaluate the three predictions from Section~\ref{sec:predictions}.
Section~\ref{sec:exp-validity} tests non-redundancy; Sections~\ref{sec:exp-readgate}--\ref{sec:exp-conditional} evaluate when \rg helps; and Sections~\ref{sec:exp-differential}--\ref{sec:exp-gemini} compare runtime constraints with reasoning-effort changes.


\subsection{Discipline and Post-Gold-Read Failures Are Non-Redundant}
\label{sec:exp-validity}

\begin{figure}[t]
    \centering
    \maybegraphic[width=\linewidth]{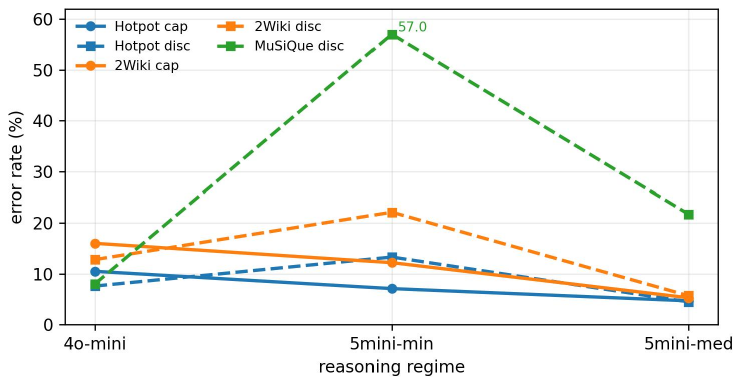}
    \caption{Error indicators across agent regimes. Discipline failures peak at gpt-5-mini minimal; post-gold-read errors follow a different curve. The x-axis is regime-level, not a scaling curve.}
    \label{fig:regime-curve}
\end{figure}

We first test whether discipline and post-gold-read failures identify distinct
trajectory regimes rather than two names for the same wrong answers. Figure~\ref{fig:regime-curve}
shows a three-regime pattern: discipline failures are lower under gpt-4o-mini
(HotpotQA 7.6\%, 2WikiMultiHopQA 12.8\%), rise under gpt-5-mini minimal
(13.3\%, 22.1\%, and 57.0\% on HotpotQA, 2WikiMultiHopQA, and MuSiQue),
and decrease under medium reasoning on HotpotQA and 2WikiMultiHopQA.

Post-gold-read errors follow a different curve on datasets with gold-evidence annotations: they generally decrease with reasoning effort, while discipline failures peak in the low-effort regime where the model can use tools but does not reliably execute the evidence-gathering procedure. 

\begin{table}[t]
\centering
\small
\caption{\textbf{Forced reading recovers skipped-read cases.}
LLM-Acc on the self-selected zero-read subset; paired McNemar $p<10^{-4}$.
This estimates a rescue effect, not the population-level marginal effect
reported in Table~\ref{tab:conditional-rg}.}
\label{tab:induced-read}
\setlength{\tabcolsep}{4pt}
\begin{tabular}{lrrrr}
\toprule
Dataset & $n$ & Zero-read & Forced read & $\Delta$ \\
\midrule
HotpotQA & 215 & 58.1 & 73.0 & \cellcolor{cellpos}$\mathbf{+14.9}$ \\
2Wiki    & 261 & 42.1 & 62.1 & \cellcolor{cellpos}$\mathbf{+19.9}$ \\
MuSiQue  & 617 & 22.5 & 37.4 & \cellcolor{cellpos}$\mathbf{+14.9}$ \\
\bottomrule
\end{tabular}
\end{table}

The strongest evidence for non-redundancy comes from a multi-label reclassification that allows discipline and post-gold-read indicators to co-occur. Among 3{,}807 wrong cases across the 12{,}000 OpenAI-family trajectories, the both-trigger rate $P(\text{disc}\wedge\text{post}\mid\text{wrong})$ stays in a narrow $[11.2\%, 13.1\%]$ band across two independent entity extractors (regex / spaCy \texttt{en\_core\_web\_sm}; the binary discipline indicator agrees at Cohen's $\kappa=0.628$, observed agreement $82.1\%$). Under both extractors the overlap sits far below the $60\%$ level at which the two axes would collapse into one, and the discipline-only share dominates the post-gold-read-only share by more than two to one (regex: 46.5\% vs 21.4\%; spaCy: 50.2\% vs 19.5\%; Appendix~\ref{app:robustness}). The small, extractor-stable overlap is the key result: many wrong answers occur before evidence-conditioned reasoning is tested, while a separate set occurs after the agent has inspected gold-supporting evidence. This full-trace accounting uses the operational indicators as implemented; post-gold-read interpretation relies on the HotpotQA and 2WikiMultiHopQA portions where complete gold evidence is available.

A strict no-read-only lower bound preserves the same minimal-reasoning discipline hump: HotpotQA 9.3\%, 2WikiMultiHopQA 15.4\%, and MuSiQue 51.5\%. Thus, the main discipline signal is not driven only by snippet-only or low-evidence heuristics. This satisfies prediction H1.

\begin{table*}[t]
\centering
\small
\caption{\textbf{Three-arm mechanism ablation.} gpt-5-mini minimal, paired $n=1{,}000$ per dataset. Deltas are LLM-Acc changes relative to no-\rg. \rg{} improves accuracy on all three datasets, whereas ctx-inject does not replicate the gain and is net-negative on 2Wiki.}
\label{tab:mechanism-ablation}
\setlength{\tabcolsep}{8pt}
\begin{tabular}{lrlll}
\toprule
Dataset  & no-\rg & \rg ($\Delta$)                 & ctx-inj ($\Delta$)                & \rg{} vs ctx $p$ \\
\midrule
HotpotQA & 79.6   & \cellcolor{rowhi}\textbf{82.8} ($\mathbf{+3.2}$) & 79.5 ($-0.1$)                          & $1.4{\times}10^{-4}$ \\
2Wiki    & 64.4   & \cellcolor{rowhi}\textbf{69.7} ($\mathbf{+5.3}$) & \cellcolor{cellneg}57.0 ($-7.4$)       & $<10^{-22}$ \\
MuSiQue  & 34.2   & \cellcolor{rowhi}\textbf{43.6} ($\mathbf{+9.4}$) & 38.1 ($+3.9$)                          & $1.0{\times}10^{-5}$ \\
\bottomrule
\end{tabular}
\end{table*}

\subsection{\rg Recovers Zero-Read Failures}
\label{sec:exp-readgate}

We next use \rg as a targeted intervention for trajectories that terminate before evidence inspection. The goal is not to make \rg a universally optimal decoding policy, but to test whether a measurable subset of errors occurs because the agent finalizes before reading. If the model emits a final answer before reading, the environment rejects the action and continues the loop. Retrieval, model weights, the judge, and reasoning effort remain unchanged.

The most direct diagnostic test is the paired induced-read contrast. We identify examples with zero-read finalization under the voluntary minimal-reasoning policy, then rerun the same questions under a forced-read environment. Table~\ref{tab:induced-read} reports a within-subset paired contrast: on the zero-read subset, forced reading raises LLM-Acc by 14.9--19.9 points across datasets, with paired McNemar tests below $10^{-4}$. This subset is self-selected by the voluntary policy and is not a random population; the contrast measures the rescue effect on cases the agent would otherwise skip rather than the marginal effect of forced reading at the population level (which is reported separately as the +3.2--9.4 full-cell \rg gain in \S\ref{sec:exp-conditional}).

A prompt-only control across the three minimal-reasoning cells reduces
zero-read behavior but does not replicate \rg's accuracy gain. In contrast,
\rg both nearly eliminates zero-read finalization and improves LLM-Acc on all
three datasets (Appendix~\ref{app:prompt-control}). Thus, instruction alone
can reduce the surface behavior, but it does not recover the accuracy gain of
an execution-level constraint.

A rank analysis shows that \rg primarily enforces that a read happens, not which chunk is read: forced reads remain rank-1 dominated, and 94\% of forced reads on gold-annotated datasets land on gold-supporting chunks (Appendix~\ref{app:rg-rank}).

On the full minimal-reasoning cells, \rg improves LLM-Acc by 3.2 points on HotpotQA, 5.3 points on 2WikiMultiHopQA, and 9.4 points on MuSiQue. Appendix~\ref{app:rg-overhead} reports the corresponding overhead profile, including reads, loop count, corrective interventions, retrieved tokens, and LLM-Acc. Appendix~\ref{app:contain-acc} reports Contain-Acc for the same cells; its direction matches LLM-Acc on all three datasets.

\subsection{\rg Helps When Residual Discipline Error Is High}
\label{sec:exp-conditional}

\rg is conditional: it helps most when residual discipline error is high. We compare discipline error before \rg against the \rg change in LLM-Acc across full \rg cells (Table~\ref{tab:conditional-rg}; visualization in Appendix~\ref{app:conditional-rg}). Minimal-reasoning cells show a monotonic pattern: larger residual discipline error corresponds to larger \rg gains. Medium-reasoning cells have lower residual discipline error and show zero or negative gains. This is expected boundary behavior rather than a contradiction. When the agent already reads reliably, the gate has little headroom and can introduce unnecessary interventions.

\begin{table}[t]
\centering
\small
\caption{\textbf{\rg's gain tracks residual discipline error.} Baseline and $\edisc$ are no-\rg{} values; $\Delta$~\rg{} is the paired LLM-Acc change. Minimal-reasoning rows recover accuracy (green); medium-reasoning rows do not (red), consistent with H2. $^{\dagger}$Medium rows use the same $n=100$ matched-question subsample as the paired \rg ablation (Appendix~\ref{app:rg-overhead}); the same no-\rg{} system's full $n=1{,}000$ cell mean is 90.8/88.8/67.1 (Appendix~\ref{app:architecture}) --- the two are consistent measurements on different question subsamples, not two values for different systems.}
\label{tab:conditional-rg}
\setlength{\tabcolsep}{4pt}
\resizebox{\linewidth}{!}{%
\begin{tabular}{llrrr}
\toprule
Effort & Dataset & $\edisc$ (\%) & Baseline & $\Delta$\,\rg \\
\midrule
\multirow{3}{*}{\textbf{Minimal}}
 & HotpotQA & \textbf{13.3} & 79.6 & \cellcolor{cellhi}$\mathbf{+3.2}$ \\
 & 2Wiki    & \textbf{22.1} & 64.4 & \cellcolor{cellhi}$\mathbf{+5.3}$ \\
 & MuSiQue  & \textbf{57.0} & 34.2 & \cellcolor{cellpos}$\mathbf{+9.4}$ \\
\midrule
\multirow{3}{*}{Medium$^{\dagger}$}
 & HotpotQA &  4.5 & 95.0 & \cellcolor{cellneu}$+0.0$ \\
 & 2Wiki    &  5.7 & 90.0 & \cellcolor{cellneg}$-7.0$ \\
 & MuSiQue  & 21.7 & 74.0 & \cellcolor{cellneg}$-4.0$ \\
\bottomrule
\end{tabular}}
\end{table}

This satisfies prediction H2: \rg's benefit tracks residual discipline error rather than reasoning level alone. Dataset difficulty is unlikely to confound this pattern: a gold-chunk retrieval-success proxy has near-zero cross-cell variance and cannot explain the spread in $\Delta$\,\rg; controlling instead for retrieval effort (total retrieved tokens), the correlation survives and strengthens (partial $r=+0.881$, $n=6$), holding directionally at $+0.56$ pooled with $n=13$ cross-family cells. Given the small cell counts we report this as a robustness check rather than a confirmatory test. The cross-family cells are Qwen2.5-Instruct (3B/7B/14B) runs, where the same dose-response direction reproduces at $r=+0.629$ (vs.\ $+0.728$ for the OpenAI family; Appendix~\ref{app:qwen-generalization}).

Under medium reasoning, \rg does not merely fail to help; it increases mean loop count on all three datasets while accuracy does not improve. Matched-question-ID paired means ($n=100$/dataset) rise from 3.92 to 4.25 on HotpotQA, 5.08 to 6.10 on 2WikiMultiHopQA, and 6.90 to 7.44 on MuSiQue, so additional trajectory steps do not explain the minimal-regime gains reported above. On 2Wiki-medium, where the effect is significant ($-7.0$, Table~\ref{tab:conditional-rg}), the max-loop rate rises from 15\% to 35\% under \rg, and all 8 paired cases where \rg breaks a previously correct answer terminate at the loop cap with an abstention or partial final answer --- forced reads exhaust the budget of an already-competent agent rather than giving it more useful steps.

\subsection{Mechanism: Not Merely Context Injection}
\label{sec:exp-mechanism}

An alternative interpretation of \rg's gain is that it merely pushes additional chunk text into the agent's context window. We test this with a three-arm paired ablation across all three datasets (gpt-5-mini minimal; paired by question ID; all arms share retrieval, backbone, loop budget, and judge). Arm (i) is no-\rg; (ii) is \rg; (iii) (ctx-inject) uses the same trigger as (ii) but silently appends the rank-1 chunk text as a \texttt{user}-role observation without issuing a \texttt{read} tool call (Table~\ref{tab:mechanism-ablation}).

Across all three datasets, ctx-inject fails to replicate \rg's gain and is
sometimes net-negative. On HotpotQA, ctx-inject changes LLM-Acc by $-0.1$~pp
while \rg improves it by $+3.2$~pp; on 2WikiMultiHopQA, ctx-inject reduces
accuracy by $-7.4$~pp while \rg improves it by $+5.3$~pp; on MuSiQue,
ctx-inject improves accuracy by $+3.9$~pp but still recovers less than half
of \rg's $+9.4$~pp gain. On every dataset, the paired \rg{} vs.\ ctx-inject
contrast is highly significant (McNemar $p \leq 10^{-4}$). Delivering the
same chunk text without the \texttt{read} action therefore recovers at most a
minority of \rg's gain and can be net-negative, so the gain is not explained
by additional context alone. A follow-up four-arm probe on MuSiQue
($n=500$, gpt-5-mini minimal) decomposes this further by separating delivery
channel from the self-issued read action: relative to no-\rg, \rg gives
$+8.2$~pp, a fabricated \texttt{tool}-role injection gives $+5.2$~pp, and a
\texttt{user}-role injection gives $+4.4$~pp, so the channel difference
($+0.8$~pp) is not significant while committing to the read action itself
adds a further $+3.0$~pp over tool-channel injection
(Appendix~\ref{app:toolrole}). This decomposes, rather than contradicts, the
three-arm result above: across these controls, the residual gain tracks the
agent's self-issued read action rather than the channel through which the
text is delivered. We treat this action-commitment account as the
interpretation best supported by the controls, not as a conclusively
isolated mechanism.

\paragraph{When to enable \rg{} in practice.}
A practical trigger for enabling \rg does not require gold labels. The most
direct proxy is the zero-read finalization rate in unlabeled logs: if an
agent often searches and then finalizes with \texttt{read\_count}$=0$, it is
a natural candidate for the gate. Gold annotations are needed to evaluate
accuracy and post-gold-read failures, but not to detect the procedural
violation itself.

\subsection{Runtime Constraints and Reasoning Effort Have Different Signatures}
\label{sec:exp-differential}

\begin{table}[t]
\centering
\small
\caption{\textbf{\rg and reasoning effort target different indicators.}
The marginal \rg$\to\ppost$ OR is exposure-induced re-classification; the
within-strata row ($n=1{,}863$, read-exposure matched) is the behavioral
estimate. Correction details are in Appendix~\ref{app:fdr}.}
\label{tab:differential-response}
\setlength{\tabcolsep}{4pt}
\resizebox{\linewidth}{!}{%
\begin{tabular}{lllrl}
\toprule
Intervention & Spec.\ & Indicator & OR & 95\% CI / test \\
\midrule
\multirow{3}{*}{\rg}
 & full           & $\pdisc$           & $\mathbf{0.38}$ & $[0.35,0.41]$ \\
 & marginal       & $\ppost$           & 1.46            & $[1.36,1.57]$ \\
 & within-strata  & $\ppost$           & $\mathbf{1.00}$ & $[0.84,1.19]$ \\
\midrule
\multirow{2}{*}{Effort}
 & med vs min     & $\pdisc$           & $\mathbf{0.22}$ & $[0.20,0.26]$ \\
 & med vs min     & $\ppost$           & 0.51            & $[0.43,0.61]$ \\
\midrule
Interaction       & effort $\times$ axis & stacked          & $\mathbf{2.02}$ & $p<10^{-4}$ \\
Perm.\ test       & ---                  & $T_{\mathrm{obs}}$ & 18.6 pp        & $p<10^{-4}$ \\
\bottomrule
\end{tabular}%
}
\end{table}

We next ask whether \rg merely mimics additional reasoning effort, or whether
it changes a different part of the agent trajectory. The results support the
second view: procedural constraints and reasoning effort produce different
signatures over $(\pdisc,\ppost)$ (Table~\ref{tab:differential-response}; see
Appendix~\ref{app:intervention-plane} for a visualization).

We fit question-clustered logistic models over the OpenAI trajectories;
\epost{} terms restrict to HotpotQA and 2WikiMultiHopQA, which carry
per-chunk gold annotations. \rg strongly reduces discipline failure
(\pdisc{} OR $0.38$ $[0.35,0.41]$), as expected for an intervention that
blocks finalization before reading. Its effect on post-gold-read failure
requires more care. Marginally, \rg appears to increase \ppost{} (OR $1.46$
$[1.36,1.57]$), but this is a re-classification effect: the gate converts
some no-read wrong answers into with-read wrong answers, making them eligible
for the post-gold-read label. Indeed, all 106 paired \ppost{} label flips are
$0\to1$ cases driven by \texttt{read\_count} changing from 0 to positive.

To estimate the behavioral effect after read exposure is held fixed, we
compare trajectories with \texttt{read\_count}$>0$ in both conditions
($n=1{,}863$). In this within-strata comparison, \rg{}'s OR for \ppost{} is
$1.00$ $[0.84,1.19]$; the stricter gold-read-both stratum
(HotpotQA+2Wiki, $n=1{,}471$) gives the same OR, $1.00$ $[0.83,1.20]$.
Thus, \rg changes whether evidence is inspected, but does not measurably
change post-read answering once read exposure is matched.

Reasoning effort has a different signature. Increasing gpt-5-mini from
minimal to medium effort with \rg{} off lowers both indicators: \pdisc{} OR
$0.22$ $[0.20,0.26]$ and \ppost{} OR $0.51$ $[0.43,0.61]$, with a significant
effort-by-axis interaction (OR $2.02$, $p<10^{-4}$). In other words, \rg
mainly repairs whether evidence is inspected, whereas additional reasoning
effort affects both inspection behavior and post-read answering. A
within-cell-stratified label-permutation test further rejects exchangeability
of \pdisc{} and \ppost{} under \rg{} ($T_{\mathrm{obs}}=18.6$ pp, null 99th
percentile $3.5$ pp, $p<10^{-4}$; Appendix~\ref{app:robustness}). This
satisfies prediction H3: \rg{} and reasoning effort produce distinguishable
signatures over $(\pdisc,\ppost)$ even after read-exposure is held fixed.

\subsection{Hidden Thinking Does Not Guarantee Evidence Inspection}
\label{sec:exp-gemini}

We use Gemini 2.5 Flash as an external diagnostic and repeat the thinking-budget contrast on three datasets. With the same two-tool loop and no \rg, enabling a 1024-token thinking budget increases zero-read finalization on all three datasets and reduces paired correctness (Table~\ref{tab:gemini-thinking}).
On HotpotQA, zero-read rises by $+5.7$~pp with a net loss of $44$ correct
answers (McNemar $p=0.0025$); on 2WikiMultiHopQA, zero-read rises by
$+24.8$~pp with a net loss of $67$ correct answers (McNemar $p=0.0001$);
on MuSiQue, zero-read rises by $+42.6$~pp with a net loss of $73$ correct
answers (McNemar $p=3.1{\times}10^{-5}$). The zero-read shift is largest on
MuSiQue and smallest on HotpotQA, suggesting that the effect may be more
pronounced when answers require longer evidence chains.
Appendix~\ref{app:gemini-example} provides additional break-type analysis.

\begin{table}[t]
\centering
\small
\caption{\textbf{Gemini 2.5 Flash thinking budget increases zero-read
finalization on all three datasets.} Paired contrast between thinking budget
off ($b=0$) and on ($b=1024$), with no \rg and matched $n=1{,}000$ question
IDs per dataset.}
\label{tab:gemini-thinking}
\setlength{\tabcolsep}{4pt}
\begin{tabular}{lrrr}
\toprule
Dataset & ZR $\Delta$ & Net $\Delta$ & McNemar $p$ \\
\midrule
HotpotQA  & $+5.7$ pp  & \cellcolor{cellneg}$-44$ & 0.0025 \\
2Wiki     & \cellcolor{cellneg}$+24.8$ pp & \cellcolor{cellneg}$-67$ & 0.0001 \\
MuSiQue   & \cellcolor{cellneg}$+42.6$ pp & \cellcolor{cellneg}$-73$ & $3.1{\times}10^{-5}$ \\
\bottomrule
\end{tabular}
\end{table}

The replication does not imply that reasoning effort is harmful in general.
In this diagnostic setting, the result supports the broader claim that
hidden deliberation and evidence inspection can vary independently: a model
can spend more compute while still failing to execute the external
evidence-gathering procedure. The larger shift on MuSiQue suggests that this
failure mode may be more pronounced when evidence chains are harder to
recover from internal memory alone.

\paragraph{Cost--accuracy implication.}
\label{sec:cost-tradeoff}
Adding \rg to gpt-5-mini at minimal reasoning closes $20$--$24\%$ of the
LLM-Acc gap to gpt-5-mini at medium reasoning, with modest overhead in
reads and loop count (Appendix~\ref{app:cost-tradeoff}). This suggests
\rg can be a partial substitute for higher reasoning effort when residual
discipline error is high, but not a replacement for stronger reasoning.
Appendix~\ref{app:architecture} reports an additional interface-granularity
comparison as supporting context.

\paragraph{Robustness and scope.}
\label{sec:exp-robustness}
Robustness checks support the same qualitative conclusions: the discipline
signal is stable under low-evidence threshold sweeps and a strict
no-read-only lower bound; paired bootstrap intervals exclude zero for the
headline \rg gains and induced-read contrast; and the label-permutation
result remains significant under a within-cell-stratified null. Full
details are in Appendix~\ref{app:robustness}. Our claims are scoped to
Wikipedia-style multi-hop QA, where snippet-grounded shortcutting and
explicit evidence chains make discipline failures visible.

\section{Conclusion}

Agentic RAG systems can fail before evidence-conditioned reasoning begins. In
Wikipedia-style multi-hop QA, we show that a substantial subset of wrong
answers arises when the agent searches, observes plausible surface evidence,
and finalizes without sufficient reading. These failures are not ordinary
post-evidence reasoning errors: the system has not yet reached the state in
which its answer is grounded in inspected evidence.

We make this bottleneck measurable through a trajectory-level decomposition
that separates discipline failures from post-gold-read failures. Across paired
trajectories, these indicators identify largely non-redundant error regimes
and respond differently to runtime constraints and reasoning-effort changes.
\rg recovers zero-read failures when residual discipline error is high, but
has little headroom when the agent already reads reliably. This supports a
diagnostic view of \rg: it is not a universally optimal control policy, but a
minimal intervention for testing whether evidence inspection itself is the
bottleneck.

\paragraph{Future work.}
Future work should test this decomposition across larger controllers,
additional model families, non-English settings, and non-Wikipedia RAG
domains, while studying whether the self-issued read action's advantage over
passively delivered context generalizes across backbones and domains.

Final-answer accuracy remains necessary, but it is incomplete without
trajectory-level measurements of evidence inspection.
\section*{Limitations}

\paragraph{Scope.}
Our main experiments focus on cost-sensitive agent controllers
(gpt-4o-mini and gpt-5-mini), with Gemini 2.5 Flash used as an external
diagnostic. This operating regime is aligned with \rg's intended use: the gate
has the most headroom when residual discipline error is high
(\S\ref{sec:exp-conditional}). Larger frontier controllers, additional model
families, and non-English settings remain important directions for future
work. The context-injection mechanism check (\S\ref{sec:exp-mechanism}) is
run on three multi-hop datasets at gpt-5-mini minimal, and the
channel-vs-action decomposition (Appendix~\ref{app:toolrole}) is run on a
single dataset and backbone; whether the self-issued-action effect
generalizes identically across other backbones and datasets is open. The
framework also presumes discrete \texttt{search}/\texttt{read}/\texttt{final}
tool actions (\S\ref{sec:framework}); agents that interleave retrieval and
generation implicitly expose no read boundary to observe or gate on.

\paragraph{Domain dependence.}
All datasets are English Wikipedia-style multi-hop QA benchmarks, where
snippet-grounded shortcuts and explicit evidence chaining make pre-evidence
failures visible. Other RAG domains have different snippet semantics and
retrieval requirements. The trajectory-level decomposition may transfer, but
entity-coverage thresholds and gate behavior should be re-calibrated per
domain.

\paragraph{Retrieval and safety.}
\rg is not a retrieval-quality fix: it assumes that search returns useful
candidates often enough that forcing a read can expose relevant evidence. It
also modestly increases reads-per-question and retrieved tokens
(Table~\ref{tab:rg-overhead}). Production systems should therefore combine
forced-read mechanisms with retrieval-quality monitoring, filtering, and
answer-side verification, especially in sensitive domains.
\section*{Ethics Statement}

This work uses public QA benchmarks and does not involve human subjects or private user data. The main risks are deployment risks common to RAG systems. A procedural gate can improve evidence inspection, but it does not guarantee factual correctness. Systems using such gates should still retain answer verification, logging, and domain-specific safety checks, especially in high-stakes settings. The results should not be used to justify deploying low-effort agents without downstream validation.

\paragraph{Use of Scientific Artifacts.}
Our experiments use open-source tools including PyTorch \citep{pytorch}, with the Qwen3-Embedding-0.6B dense retriever \citep{qwen3embedding} obtained via the HuggingFace \citep{huggingface} library. For agent controllers and LLM-as-judge scoring we use OpenAI's API (gpt-4o-mini, gpt-5-mini) under their sharing and publication policy, and Google's Gemini API (Gemini 2.5 Flash as an external thinking-budget diagnostic and Gemini 2.5 Pro for cross-judge robustness) under their respective terms of service. All datasets (HotpotQA, 2WikiMultiHopQA, MuSiQue) are publicly released by their authors under their original licenses and used as intended for research evaluation.

\paragraph{Use of AI Assistants.}
We used Google Gemini to refine wording and polish prose. Some experimental code was written with assistance from Anthropic Claude. All final scientific claims, experimental designs, and analyses are the authors' own; AI assistants were used only as a writing and coding aid, and no AI-generated content was incorporated without author review and verification.


\bibliography{references}

\begin{thebibliography}{40}
\providecommand{\natexlab}[1]{#1}

\bibitem[{Asai et~al.(2024)Asai, Wu, Wang, Sil, and Hajishirzi}]{asai2024selfrag}
Akari Asai, Zeqiu Wu, Yizhong Wang, Avirup Sil, and Hannaneh Hajishirzi. 2024.
\newblock {Self-RAG}: Learning to retrieve, generate, and critique through self-reflection.
\newblock In \emph{International Conference on Learning Representations}.

\bibitem[{Chen et~al.(2025)Chen, Sun, Li, Sun, Zhou, Zhu, Wang, Pan, Zhang, Chen, Yang, Zhou, and Chen}]{chen2025research}
Mingyang Chen, Linzhuang Sun, Tianpeng Li, Haoze Sun, Yijie Zhou, Chenzheng Zhu, Haofen Wang, Jeff~Z. Pan, Wen Zhang, Huajun Chen, Fan Yang, Zenan Zhou, and Weipeng Chen. 2025.
\newblock {ReSearch}: Learning to reason with search for {LLMs} via reinforcement learning.
\newblock \emph{arXiv preprint arXiv:2503.19470}.

\bibitem[{Dhuliawala et~al.(2024)Dhuliawala, Komeili, Xu, Raileanu, Li, Celikyilmaz, and Weston}]{dhuliawala2024chain}
Shehzaad Dhuliawala, Mojtaba Komeili, Jing Xu, Roberta Raileanu, Xian Li, Asli Celikyilmaz, and Jason Weston. 2024.
\newblock Chain-of-verification reduces hallucination in large language models.
\newblock In \emph{Findings of the Association for Computational Linguistics: ACL 2024}.

\bibitem[{Du et~al.(2026)Du, Xu, Zhu, Wang, Wang, Wang, and Mao}]{du2026arag}
Mingxuan Du, Benfeng Xu, Chiwei Zhu, Shaohan Wang, Pengyu Wang, Xiaorui Wang, and Zhendong Mao. 2026.
\newblock \href {https://arxiv.org/abs/2602.03442} {{A-RAG}: Scaling agentic retrieval-augmented generation via hierarchical retrieval interfaces}.
\newblock \emph{Preprint}, arXiv:2602.03442.

\bibitem[{Es et~al.(2024)Es, James, Espinosa-Anke, and Schockaert}]{es2024ragas}
Shahul Es, Jithin James, Luis Espinosa-Anke, and Steven Schockaert. 2024.
\newblock {RAGAS}: Automated evaluation of retrieval augmented generation.
\newblock In \emph{Proceedings of the 18th Conference of the European Chapter of the Association for Computational Linguistics: System Demonstrations}, pages 150--158.

\bibitem[{Gao et~al.(2023)Gao, Yen, Yu, and Chen}]{gao2023alce}
Tianyu Gao, Howard Yen, Jiatong Yu, and Danqi Chen. 2023.
\newblock Enabling large language models to generate text with citations.
\newblock In \emph{Proceedings of the 2023 Conference on Empirical Methods in Natural Language Processing}, pages 6465--6488.

\bibitem[{Ho et~al.(2020)Ho, Nguyen, Sugawara, and Aizawa}]{ho2020constructing}
Xanh Ho, Anh-Khoa~Duong Nguyen, Saku Sugawara, and Akiko Aizawa. 2020.
\newblock Constructing a multi-hop {QA} dataset for comprehensive evaluation of reasoning steps.
\newblock In \emph{Proceedings of the 28th International Conference on Computational Linguistics}, pages 6609--6625.

\bibitem[{Jiang et~al.(2023)Jiang, Xu, Gao, Sun, Liu, Dwivedi-Yu, Yang, Callan, and Neubig}]{jiang2023active}
Zhengbao Jiang, Frank~F. Xu, Luyu Gao, Zhiqing Sun, Qian Liu, Jane Dwivedi-Yu, Yiming Yang, Jamie Callan, and Graham Neubig. 2023.
\newblock Active retrieval augmented generation.
\newblock In \emph{Proceedings of the 2023 Conference on Empirical Methods in Natural Language Processing}, pages 7969--7992.

\bibitem[{Jin et~al.(2025)Jin, Zeng, Yue, Yoon, Arik, Wang, Zamani, and Han}]{jin2025searchr1}
Bowen Jin, Hansi Zeng, Zhenrui Yue, Jinsung Yoon, Sercan Arik, Dong Wang, Hamed Zamani, and Jiawei Han. 2025.
\newblock {Search-R1}: Training {LLMs} to reason and leverage search engines with reinforcement learning.
\newblock \emph{arXiv preprint arXiv:2503.09516}.

\bibitem[{Kadavath et~al.(2022)Kadavath, Conerly, Askell, Henighan, Drain, Perez, Schiefer, Hatfield-Dodds, DasSarma, Tran-Johnson, Johnston, El-Showk, Jones, Elhage, Hume, Chen, Bai, Bowman, Fort, Ganguli, Hernandez, Jacobson, Kernion, Kravec, Lovitt, Ndousse, Olsson, Ringer, Amodei, Brown, Clark, Joseph, Mann, McCandlish, Olah, and Kaplan}]{kadavath2022know}
Saurav Kadavath, Tom Conerly, Amanda Askell, Tom Henighan, Dawn Drain, Ethan Perez, Nicholas Schiefer, Zac Hatfield-Dodds, Nova DasSarma, Eli Tran-Johnson, Scott Johnston, Sheer El-Showk, Andy Jones, Nelson Elhage, Tristan Hume, Anna Chen, Yuntao Bai, Sam Bowman, Stanislav Fort, Deep Ganguli, Danny Hernandez, Josh Jacobson, Jackson Kernion, Shauna Kravec, Liane Lovitt, Kamal Ndousse, Catherine Olsson, Sam Ringer, Dario Amodei, Tom Brown, Jack Clark, Nicholas Joseph, Ben Mann, Sam McCandlish, Chris Olah, and Jared Kaplan. 2022.
\newblock Language models (mostly) know what they know.
\newblock \emph{arXiv preprint arXiv:2207.05221}.

\bibitem[{Kojima et~al.(2022)Kojima, Gu, Reid, Matsuo, and Iwasawa}]{kojima2022large}
Takeshi Kojima, Shixiang~Shane Gu, Machel Reid, Yutaka Matsuo, and Yusuke Iwasawa. 2022.
\newblock Large language models are zero-shot reasoners.
\newblock In \emph{Advances in Neural Information Processing Systems}.

\bibitem[{Kuhn et~al.(2023)Kuhn, Gal, and Farquhar}]{kuhn2023semantic}
Lorenz Kuhn, Yarin Gal, and Sebastian Farquhar. 2023.
\newblock Semantic uncertainty: Linguistic invariances for uncertainty estimation in natural language generation.
\newblock In \emph{International Conference on Learning Representations}.

\bibitem[{Lewis et~al.(2020)Lewis, Perez, Piktus, Petroni, Karpukhin, Goyal, K{\"u}ttler, Lewis, Yih, Rockt{\"a}schel, Riedel, and Kiela}]{lewis2020retrieval}
Patrick Lewis, Ethan Perez, Aleksandra Piktus, Fabio Petroni, Vladimir Karpukhin, Naman Goyal, Heinrich K{\"u}ttler, Mike Lewis, Wen-tau Yih, Tim Rockt{\"a}schel, Sebastian Riedel, and Douwe Kiela. 2020.
\newblock Retrieval-augmented generation for knowledge-intensive nlp tasks.
\newblock In \emph{Advances in Neural Information Processing Systems}.

\bibitem[{Lightman et~al.(2023)Lightman, Kosaraju, Burda, Edwards, Baker, Lee, Leike, Schulman, Sutskever, and Cobbe}]{lightman2023lets}
Hunter Lightman, Vineet Kosaraju, Yura Burda, Harri Edwards, Bowen Baker, Teddy Lee, Jan Leike, John Schulman, Ilya Sutskever, and Karl Cobbe. 2023.
\newblock Let's verify step by step.
\newblock \emph{arXiv preprint arXiv:2305.20050}.

\bibitem[{Lin et~al.(2022)Lin, Hilton, and Evans}]{lin2022teaching}
Stephanie Lin, Jacob Hilton, and Owain Evans. 2022.
\newblock Teaching models to express their uncertainty in words.
\newblock \emph{Transactions on Machine Learning Research}.

\bibitem[{Liu et~al.(2023{\natexlab{a}})Liu, Zhang, and Liang}]{liu2023verifiability}
Nelson~F. Liu, Tianyi Zhang, and Percy Liang. 2023{\natexlab{a}}.
\newblock Evaluating verifiability in generative search engines.
\newblock In \emph{Findings of the Association for Computational Linguistics: EMNLP 2023}.

\bibitem[{Liu et~al.(2024)Liu, Yu, Zhang, Xu, Lei, Lai, Gu, Ding, Men, Yang, Zhang, Deng, Zeng, Du, Zhang, Shen, Zhang, Su, Sun, Huang, Dong, and Tang}]{liu2023agentbench}
Xiao Liu, Hao Yu, Hanchen Zhang, Yifan Xu, Xuanyu Lei, Hanyu Lai, Yu~Gu, Hangliang Ding, Kaiwen Men, Kejuan Yang, Shudan Zhang, Xiang Deng, Aohan Zeng, Zhengxiao Du, Chenhui Zhang, Sheng Shen, Tianjun Zhang, Yu~Su, Huan Sun, Minlie Huang, Yuxiao Dong, and Jie Tang. 2024.
\newblock {AgentBench}: Evaluating {LLMs} as agents.
\newblock In \emph{International Conference on Learning Representations}.

\bibitem[{Liu et~al.(2023{\natexlab{b}})Liu, Iter, Xu, Wang, Xu, and Zhu}]{liu2023geval}
Yang Liu, Dan Iter, Yichong Xu, Shuohang Wang, Ruochen Xu, and Chenguang Zhu. 2023{\natexlab{b}}.
\newblock {G-Eval}: {NLG} evaluation using {GPT-4} with better human alignment.
\newblock In \emph{Proceedings of the 2023 Conference on Empirical Methods in Natural Language Processing}, pages 2511--2522.

\bibitem[{Mallen et~al.(2023)Mallen, Asai, Zhong, Das, Khashabi, and Hajishirzi}]{mallen2023whennot}
Alex Mallen, Akari Asai, Victor Zhong, Rajarshi Das, Daniel Khashabi, and Hannaneh Hajishirzi. 2023.
\newblock When not to trust language models: Investigating effectiveness of parametric and non-parametric memories.
\newblock In \emph{Proceedings of the 61st Annual Meeting of the Association for Computational Linguistics}, pages 9802--9822.

\bibitem[{Nakano et~al.(2021)Nakano, Hilton, Balaji, Wu, Ouyang, Kim, Hesse, Jain, Kosaraju, Saunders, Jiang, Cobbe, Eloundou, Krueger, Button, Knight, Chess, and Schulman}]{nakano2021webgpt}
Reiichiro Nakano, Jacob Hilton, Suchir Balaji, Jeff Wu, Long Ouyang, Christina Kim, Christopher Hesse, Shantanu Jain, Vineet Kosaraju, William Saunders, Xu~Jiang, Karl Cobbe, Tyna Eloundou, Gretchen Krueger, Kevin Button, Matthew Knight, Benjamin Chess, and John Schulman. 2021.
\newblock {WebGPT}: Browser-assisted question-answering with human feedback.
\newblock \emph{arXiv preprint arXiv:2112.09332}.

\bibitem[{Paszke et~al.(2019)Paszke, Gross, Massa, Lerer, Bradbury, Chanan, Killeen, Lin, Gimelshein, Antiga, Desmaison, K{\"o}pf, Yang, DeVito, Raison, Tejani, Chilamkurthy, Steiner, Fang, Bai, and Chintala}]{pytorch}
Adam Paszke, Sam Gross, Francisco Massa, Adam Lerer, James Bradbury, Gregory Chanan, Trevor Killeen, Zeming Lin, Natalia Gimelshein, Luca Antiga, Alban Desmaison, Andreas K{\"o}pf, Edward Yang, Zachary DeVito, Martin Raison, Alykhan Tejani, Sasank Chilamkurthy, Benoit Steiner, Lu~Fang, Junjie Bai, and Soumith Chintala. 2019.
\newblock {PyTorch}: An imperative style, high-performance deep learning library.
\newblock In \emph{Advances in Neural Information Processing Systems}.

\bibitem[{Press et~al.(2023)Press, Zhang, Min, Schmidt, Smith, and Lewis}]{press2023selfask}
Ofir Press, Muru Zhang, Sewon Min, Ludwig Schmidt, Noah~A. Smith, and Mike Lewis. 2023.
\newblock Measuring and narrowing the compositionality gap in language models.
\newblock In \emph{Findings of the Association for Computational Linguistics: EMNLP 2023}, pages 5687--5711.

\bibitem[{Qin et~al.(2024)Qin, Liang, Ye, Zhu, Yan, Lu, Lin, Cong, Tang, Qian, Zhao, Hong, Tian, Xie, Zhou, Gerstein, Li, Liu, and Sun}]{qin2023toolllm}
Yujia Qin, Shihao Liang, Yining Ye, Kunlun Zhu, Lan Yan, Yaxi Lu, Yankai Lin, Xin Cong, Xiangru Tang, Bill Qian, Sihan Zhao, Lauren Hong, Runchu Tian, Ruobing Xie, Jie Zhou, Mark Gerstein, Dahai Li, Zhiyuan Liu, and Maosong Sun. 2024.
\newblock {ToolLLM}: Facilitating large language models to master 16000+ real-world {APIs}.
\newblock In \emph{International Conference on Learning Representations}.

\bibitem[{{Qwen Team}(2024)}]{qwen2.5}
{Qwen Team}. 2024.
\newblock Qwen2.5 technical report.
\newblock \emph{arXiv preprint arXiv:2412.15115}.

\bibitem[{Rashkin et~al.(2023)Rashkin, Nikolaev, Lamm, Aroyo, Collins, Das, Petrov, Singh~Tomar, Turc, and Reitter}]{rashkin2023measuring}
Hannah Rashkin, Vitaly Nikolaev, Matthew Lamm, Lora Aroyo, Michael Collins, Dipanjan Das, Slav Petrov, Gaurav Singh~Tomar, Iulia Turc, and David Reitter. 2023.
\newblock Measuring attribution in natural language generation models.
\newblock \emph{Computational Linguistics}, 49(4):777--840.

\bibitem[{Saad-Falcon et~al.(2024)Saad-Falcon, Khattab, Potts, and Zaharia}]{saadfalcon2024ares}
Jon Saad-Falcon, Omar Khattab, Christopher Potts, and Matei Zaharia. 2024.
\newblock {ARES}: An automated evaluation framework for retrieval-augmented generation systems.
\newblock In \emph{Proceedings of the 2024 Conference of the North American Chapter of the Association for Computational Linguistics}.

\bibitem[{Schick et~al.(2023)Schick, Dwivedi-Yu, Dess{\`i}, Raileanu, Lomeli, Hambro, Zettlemoyer, Cancedda, and Scialom}]{schick2023toolformer}
Timo Schick, Jane Dwivedi-Yu, Roberto Dess{\`i}, Roberta Raileanu, Maria Lomeli, Eric Hambro, Luke Zettlemoyer, Nicola Cancedda, and Thomas Scialom. 2023.
\newblock Toolformer: Language models can teach themselves to use tools.
\newblock In \emph{Advances in Neural Information Processing Systems}.

\bibitem[{Song et~al.(2025)Song, Jiang, Min, Chen, Chen, Zhao, Fang, and Wen}]{song2025r1searcher}
Huatong Song, Jinhao Jiang, Yingqian Min, Jie Chen, Zhipeng Chen, Wayne~Xin Zhao, Lei Fang, and Ji-Rong Wen. 2025.
\newblock {R1-Searcher}: Incentivizing the search capability in {LLMs} via reinforcement learning.
\newblock \emph{arXiv preprint arXiv:2503.05592}.

\bibitem[{Trivedi et~al.(2022)Trivedi, Balasubramanian, Khot, and Sabharwal}]{trivedi2022musique}
Harsh Trivedi, Niranjan Balasubramanian, Tushar Khot, and Ashish Sabharwal. 2022.
\newblock {MuSiQue}: Multihop questions via single-hop question composition.
\newblock \emph{Transactions of the Association for Computational Linguistics}, 10:539--554.

\bibitem[{Trivedi et~al.(2023)Trivedi, Balasubramanian, Khot, and Sabharwal}]{trivedi2023ircot}
Harsh Trivedi, Niranjan Balasubramanian, Tushar Khot, and Ashish Sabharwal. 2023.
\newblock Interleaving retrieval with chain-of-thought reasoning for knowledge-intensive multi-step questions.
\newblock In \emph{Proceedings of the 61st Annual Meeting of the Association for Computational Linguistics}, pages 10014--10037.

\bibitem[{Wang et~al.(2023)Wang, Wei, Schuurmans, Le, Chi, Narang, Chowdhery, and Zhou}]{wang2023selfconsistency}
Xuezhi Wang, Jason Wei, Dale Schuurmans, Quoc Le, Ed~H. Chi, Sharan Narang, Aakanksha Chowdhery, and Denny Zhou. 2023.
\newblock Self-consistency improves chain of thought reasoning in language models.
\newblock In \emph{International Conference on Learning Representations}.

\bibitem[{Wang et~al.(2025)Wang, An, Zheng, Qian, Zhang, Ouyang, Cai, Wang, and Wu}]{wang2025erase}
Ziliang Wang, Kang An, Xuhui Zheng, Faqiang Qian, Weikun Zhang, Cijun Ouyang, Jialu Cai, Yuhang Wang, and Yichao Wu. 2025.
\newblock Erase to improve: Erasable reinforcement learning for search-augmented {LLMs}.
\newblock \emph{arXiv preprint arXiv:2510.00861}.

\bibitem[{Wei et~al.(2022)Wei, Wang, Schuurmans, Bosma, Ichter, Xia, Chi, Le, and Zhou}]{wei2022chain}
Jason Wei, Xuezhi Wang, Dale Schuurmans, Maarten Bosma, Brian Ichter, Fei Xia, Ed~H. Chi, Quoc~V. Le, and Denny Zhou. 2022.
\newblock Chain-of-thought prompting elicits reasoning in large language models.
\newblock In \emph{Advances in Neural Information Processing Systems}.

\bibitem[{Wolf et~al.(2020)Wolf, Debut, Sanh, Chaumond, Delangue, Moi, Cistac, Rault, Louf, Funtowicz, Davison, Shleifer, von Platen, Ma, Jernite, Plu, Xu, {Le Scao}, Gugger, Drame, Lhoest, and Rush}]{huggingface}
Thomas Wolf, Lysandre Debut, Victor Sanh, Julien Chaumond, Clement Delangue, Anthony Moi, Pierric Cistac, Tim Rault, R{\'e}mi Louf, Morgan Funtowicz, Joe Davison, Sam Shleifer, Patrick von Platen, Clara Ma, Yacine Jernite, Julien Plu, Canwen Xu, Teven {Le Scao}, Sylvain Gugger, Mariama Drame, Quentin Lhoest, and Alexander~M. Rush. 2020.
\newblock Transformers: State-of-the-art natural language processing.
\newblock In \emph{Proceedings of the 2020 Conference on Empirical Methods in Natural Language Processing: System Demonstrations}, pages 38--45.

\bibitem[{Yan et~al.(2024)Yan, Gu, Zhu, and Ling}]{yan2024crag}
Shi-Qi Yan, Jia-Chen Gu, Yun Zhu, and Zhen-Hua Ling. 2024.
\newblock Corrective retrieval augmented generation.
\newblock In \emph{First Conference on Language Modeling}.

\bibitem[{Yang et~al.(2018)Yang, Qi, Zhang, Bengio, Cohen, Salakhutdinov, and Manning}]{yang2018hotpotqa}
Zhilin Yang, Peng Qi, Saizheng Zhang, Yoshua Bengio, William~W. Cohen, Ruslan Salakhutdinov, and Christopher~D. Manning. 2018.
\newblock {HotpotQA}: A dataset for diverse, explainable multi-hop question answering.
\newblock In \emph{Proceedings of the 2018 Conference on Empirical Methods in Natural Language Processing}, pages 2369--2380.

\bibitem[{Yao et~al.(2023)Yao, Zhao, Yu, Du, Shafran, Narasimhan, and Cao}]{yao2023react}
Shunyu Yao, Jeffrey Zhao, Dian Yu, Nan Du, Izhak Shafran, Karthik Narasimhan, and Yuan Cao. 2023.
\newblock {ReAct}: Synergizing reasoning and acting in language models.
\newblock In \emph{International Conference on Learning Representations}.

\bibitem[{Yoran et~al.(2024)Yoran, Wolfson, Ram, and Berant}]{yoran2024making}
Ori Yoran, Tomer Wolfson, Ori Ram, and Jonathan Berant. 2024.
\newblock Making retrieval-augmented language models robust to irrelevant context.
\newblock In \emph{International Conference on Learning Representations}.

\bibitem[{Zhang et~al.(2025)Zhang, Li, Long, Zhang, Lin, Yang, Xie, Yang, Liu, Lin, Huang, and Zhou}]{qwen3embedding}
Yanzhao Zhang, Mingxin Li, Dingkun Long, Xin Zhang, Huan Lin, Baosong Yang, Pengjun Xie, An~Yang, Dayiheng Liu, Junyang Lin, Fei Huang, and Jingren Zhou. 2025.
\newblock Qwen3 embedding: Advancing text embedding and reranking through foundation models.
\newblock \emph{arXiv preprint arXiv:2506.05176}.

\bibitem[{Zheng et~al.(2023)Zheng, Chiang, Sheng, Zhuang, Wu, Zhuang, Lin, Li, Li, Xing, Zhang, Gonzalez, and Stoica}]{zheng2023judging}
Lianmin Zheng, Wei-Lin Chiang, Ying Sheng, Siyuan Zhuang, Zhanghao Wu, Yonghao Zhuang, Zi~Lin, Zhuohan Li, Dacheng Li, Eric~P. Xing, Hao Zhang, Joseph~E. Gonzalez, and Ion Stoica. 2023.
\newblock Judging {LLM}-as-a-judge with {MT}-bench and chatbot arena.
\newblock In \emph{Advances in Neural Information Processing Systems, Datasets and Benchmarks Track}.

\end{thebibliography}

\newpage
\appendix

\section{Prompts and Tool Definitions}
\label{app:prompts}

For reproducibility, this appendix records the verbatim agent system prompt, the \rg corrective hints, the tool schemas, and the judge prompt used in all experiments. Question wording and gold annotations are inherited from each dataset's release.

\subsection{Agent System Prompt}
\label{app:agent-prompt}
The following prompt is used for both the no-\rg and the \rg conditions on all main cells:

\begin{promptbox}
You are a question-answering assistant with access to a corpus through hybrid retrieval tools.

\#\# Available Tools

- \textbf{search(query, k)}: Hybrid (BM25 + dense) retrieval. Returns top-k chunk IDs with the 5 most query-relevant sentences as snippets. Snippets are NOT the full chunk --- use `read' for full context. Snippet overlap across calls signals the index has converged on this query intent; reword or pivot.

- \textbf{read(evidence\_id)}: Read the full content of a specific chunk.

\#\# Strategy

Work iteratively: search $\rightarrow$ read $\rightarrow$ evaluate $\rightarrow$ search $\rightarrow$ read $\rightarrow$ \ldots $\rightarrow$ answer.

\textbf{Never re-issue a near-duplicate query.} If a search returned poor or repeated results, do ONE of: (a) reword with different surface forms, (b) add a discriminating entity/date/number from prior results, (c) decompose into a narrower sub-question, (d) stop searching and answer from what you have.

\textbf{Multi-hop discipline.} If the question chains two facts, do NOT search the full composed question. Use a hop-1 search to identify the bridge entity, read, extract entity, then hop-2 search using the bridge entity. Single-hop questions should be answered in one search+read whenever possible.

\#\# When Answering

- Before answering, you MUST `read' at least one chunk whose snippet directly supports the answer. Snippets alone are insufficient evidence. Exception: if the snippet contains the literal extractive answer span and matches the question entity exactly, you may answer without `read'.
- Ground your response in the retrieved chunks. Cite the specific chunks that support your answer.
- \textbf{Stop when:} you have read evidence that uniquely answers the question, OR two reformulations in a row produced no new useful evidence.
\end{promptbox}

The strict-prompt control replaces the snippet exception with: \textit{``Before answering, you MUST `read' at least one chunk whose text directly supports the answer. Snippets alone are never sufficient evidence. Do not answer until at least one read(evidence\_id) call has succeeded.''}

\subsection{Read-Gate Corrective Hints}
\label{app:rg-hint}
The main \rg condition used throughout the paper triggers a single environment-level check: the agent emits a \texttt{final} action with \texttt{read\_count}=0. The offending action is rejected, the read-before-final hint below is appended to the next observation, and the loop resumes without modifying any other prompt component. We additionally implement a stricter hint that triggers when the agent issues an action other than \texttt{read} immediately after a \texttt{search} call without inspecting any returned chunk (read-after-search); this hint is not part of the main \rg condition and is not separately ablated with its own reported numbers here, but it is available within the same corrective-hint framework that motivates the broader family of stronger procedural gates evaluated in Appendix~\ref{app:gate-family}.

\paragraph{Read-before-final hint (main \rg).}
\begin{promptbox}
You attempted to answer before reading any evidence. You MUST call read(evidence\_id) on a promising chunk before giving the final answer.
\end{promptbox}

\paragraph{Read-after-search hint (implemented, not used in main \rg).}
\begin{promptbox}
Your last search returned candidate chunks but you have not read any of them yet. You MUST call read(evidence\_id) on at least one of the most recently returned chunk IDs before answering or issuing another search.
\end{promptbox}

\subsection{Tool Schemas}
\label{app:tool-schemas}
The two-tool agent uses the following JSON tool schemas. \texttt{search} performs reciprocal-rank fusion over BM25 and a 0.6B dense retriever with top-$k=5$ returned to the agent and an internal fusion $k=60$. \texttt{read} returns the full chunk text by ID.

\begin{promptbox}
\{ "name": "search", "parameters": \{ "query": string, "k": integer (default 5) \} \}\\
\{ "name": "read",   "parameters": \{ "evidence\_id": string \} \}
\end{promptbox}

\subsection{Judge Prompt}
\label{app:judge-prompt}
We score answers with a fixed gpt-5-mini judge at temperature 0.0. The judge receives the question, gold reference, and predicted answer, and returns a single JSON object \texttt{\{"correct": bool, "rationale": str\}}. The judge instruction is:

\begin{promptbox}
You are a strict evaluator of question-answering systems. Given a question, a system-predicted answer, and a reference answer, decide whether the prediction conveys the same factual answer as the reference. Reply with a single JSON object: \{"correct": true$|$false, "rationale": "..."\}. Be conservative: numeric/extractive answers require exact factual match; abstractive answers require all key facts to be present. Minor wording differences are OK; missing facts or different facts are NOT OK.
\end{promptbox}

The judge runs on the question + gold + predicted-answer triple only and does not see retrieved chunks, system identity, or run metadata.

\section{Read-Gate Operational Details}
\label{app:rg-operational}

\subsection{Read-Gate Overhead}
\label{app:rg-overhead}
Table~\ref{tab:rg-overhead} reports the observable overhead of \rg on the full minimal-reasoning cells. Corrections/Q counts rejected actions that consumed a loop without an executed tool call, after subtracting the final answer turn; this inferred count is zero for all matched no-gate cells.

\begin{table*}[!t]
\centering
\small
\caption{Deployment-relevant overhead for the main gpt-5-mini minimal cells. Ret.\ toks/Q counts retrieved evidence tokens rather than total API tokens. Corrections/Q is inferred from loop turns that correspond to rejected actions rather than executed tools.}
\label{tab:rg-overhead}
\begin{tabular}{llrrrrr}
\toprule
Dataset & Condition & Reads/Q & Loops/Q & Corrections/Q & Ret.\ toks/Q & LLM-Acc \\
\midrule
HotpotQA & No gate & 0.81 & 3.07 & 0.00 & 8{,}227  & 79.6 \\
HotpotQA & \rg     & 1.03 & 3.63 & 0.25 & 9{,}011  & 82.8 \\
2Wiki    & No gate & 0.77 & 3.54 & 0.00 & 12{,}721 & 64.4 \\
2Wiki    & \rg     & 1.06 & 4.32 & 0.29 & 14{,}783 & 69.7 \\
MuSiQue  & No gate & 0.35 & 2.90 & 0.00 & 9{,}711  & 34.2 \\
MuSiQue  & \rg     & 1.00 & 4.97 & 0.89 & 14{,}192 & 43.6 \\
\bottomrule
\end{tabular}
\end{table*}

\subsection{Read-Chunk Rank Distribution Under \rg}
\label{app:rg-rank}
\begin{figure}[!htbp]
  \centering
  \maybegraphic[width=0.7\linewidth]{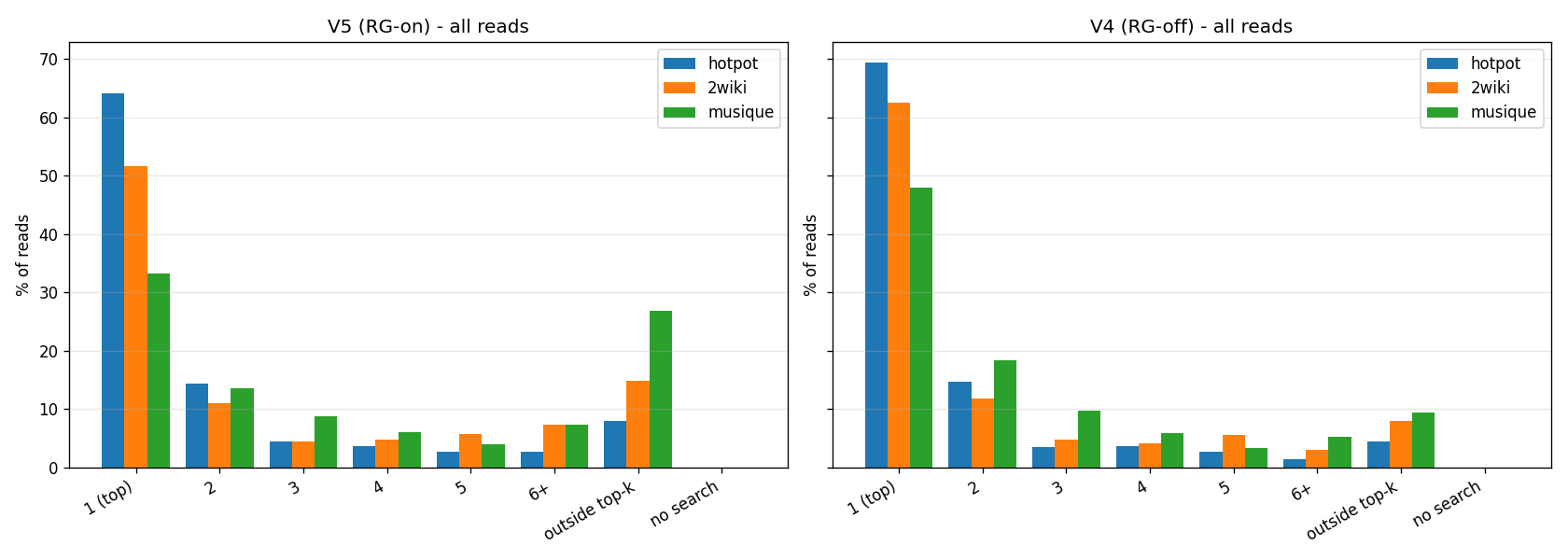}
  \caption{Distribution of read-chunk rank under \rg{} (V5) and voluntary policy (V4), gpt-5-mini minimal backbone. Rank-1 dominance is largely backbone behavior, not gate-induced.}
  \label{fig:r12-rank}
\end{figure}

Figure~\ref{fig:r12-rank} reports the rank distribution of read events under the \rg{} (V5) condition and under the voluntary policy (V4) on the same gpt-5-mini minimal backbone. Both conditions are dominated by rank-1 reads, supporting the \S\ref{sec:exp-readgate} claim that \rg{} enforces \emph{that} a read happens rather than \emph{which} chunk gets read.

Quantitatively, when \rg{} forces a read after blocking a finalization attempt, the agent picks the rank-1 retrieval result in $50.0\%$ of terminal-read events and a top-2 chunk in roughly $64\%$; $94\%$ of forced reads on gold-annotated datasets (HotpotQA + 2WikiMultiHopQA) land on a gold-supporting chunk. Without \rg, the rank-1 share is $60.0\%$ on the same backbone --- comparable, slightly higher. The agent's chunk choice is therefore already rank-1-dominated and is not changed materially by the gate.

\subsection{Contain-Acc Check}
\label{app:contain-acc}
Table~\ref{tab:contain-acc} reports Contain-Acc alongside LLM-Acc for the main minimal-reasoning cells. Contain-Acc is a string-containment secondary metric and is less semantically flexible than the LLM judge, but it moves in the same direction as LLM-Acc under \rg on all three datasets.

\begin{table}[H]
\centering
\small
\caption{LLM-Acc and Contain-Acc for the main gpt-5-mini minimal cells; \rg rows bolded.}
\label{tab:contain-acc}
\setlength{\tabcolsep}{4pt}
\begin{tabular}{llrr}
\toprule
Dataset  & Condition & LLM-Acc & Contain-Acc \\
\midrule
HotpotQA & No gate            & 79.6 & 76.8 \\
HotpotQA & \textbf{\rg}       & \textbf{82.8} & \textbf{80.1} \\
2Wiki    & No gate            & 64.4 & 72.3 \\
2Wiki    & \textbf{\rg}       & \textbf{69.7} & \textbf{75.9} \\
MuSiQue  & No gate            & 34.2 & 32.8 \\
MuSiQue  & \textbf{\rg}       & \textbf{43.6} & \textbf{41.3} \\
\bottomrule
\end{tabular}
\end{table}

\subsection{Conditional Read-Gate Benefit}
\label{app:conditional-rg}
\begin{figure}[H]
    \centering
    \maybegraphic[width=\linewidth]{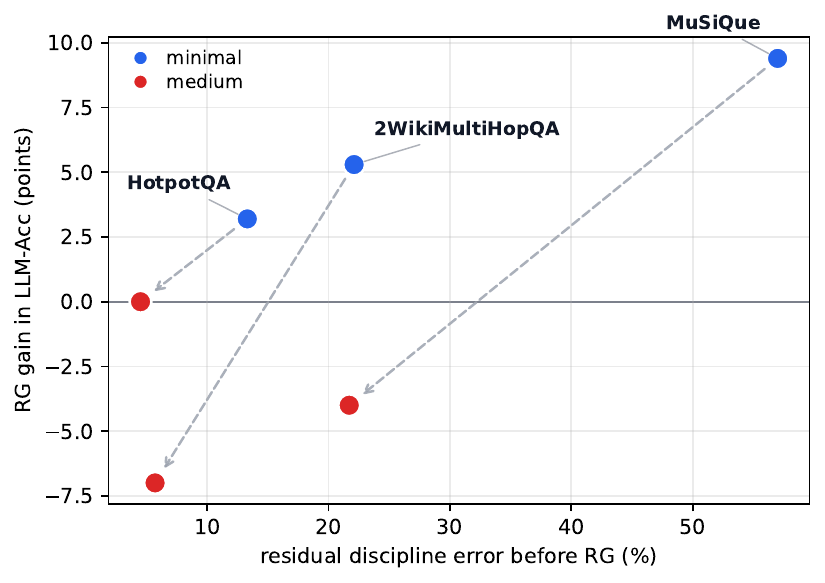}
    \caption{Conditional benefit of \rg. Each dataset contributes minimal- and medium-reasoning cells connected by a dashed arrow; gains grow with residual discipline error. Medium cells use $n=100$ matched ablations (boundary evidence only).}
    \label{fig:rg-conditional}
\end{figure}

Figure~\ref{fig:rg-conditional} visualizes the conditional benefit of \rg{} corresponding to Table~\ref{tab:conditional-rg} in the main text: across datasets, \rg's accuracy gain grows with residual discipline error.

\subsection{Cost--Accuracy Trade-off}
\label{app:cost-tradeoff}
Table~\ref{tab:cost-tradeoff} reports the cost--accuracy trade-off summarized in \S\ref{sec:cost-tradeoff}: adding \rg{} to gpt-5-mini at minimal reasoning closes $20$--$24\%$ of the LLM-Acc gap to gpt-5-mini at medium reasoning, without changing model weights or the reasoning budget. The corresponding read/loop overhead profile is in Table~\ref{tab:rg-overhead}.

\begin{table}[H]
\centering
\small
\caption{\textbf{\rg vs.\ medium reasoning within the two-tool architecture} (within-architecture trade-off; not vs.\ best system overall). Adding \rg{} at minimal reasoning (shaded row) closes $20$--$24\%$ of the LLM-Acc gap. $^{\dagger}$~Medium-reasoning rows are $n=100$ matched ablations paired with the \rg{} medium ablation; others $n=1{,}000$. The same no-\rg{} system's full $n=1{,}000$ cell mean is 90.8/88.8/67.1 (Appendix~\ref{app:architecture}), consistent with this $n=100$ subsample.}
\label{tab:cost-tradeoff}
\setlength{\tabcolsep}{4pt}
\resizebox{\linewidth}{!}{%
\begin{tabular}{lrrr}
\toprule
Configuration & HotpotQA & 2Wiki & MuSiQue \\
\midrule
5-mini-min                & 79.6 & 64.4 & 34.2 \\
\rowcolor{rowhi}\textbf{5-mini-min + \rg} & \textbf{82.8} & \textbf{69.7} & \textbf{43.6} \\
\quad gap-closing \%      & 20.8 & 20.7 & 23.6 \\
5-mini-med (target)$^{\dagger}$ & 95.0 & 90.0 & 74.0 \\
\bottomrule
\end{tabular}%
}
\end{table}

\section{Mechanism Ablation Details}
\label{app:r6-ablation}

Table~\ref{tab:r6-overhead} reports the per-arm overhead of the three-arm mechanism ablation summarized in \S\ref{sec:exp-mechanism}, across all three datasets (seed=42). All three arms use gpt-5-mini at minimal reasoning effort, gpt-5-mini judge, max\_loops=10, and max\_token\_budget=128k. The error rate (records with an \texttt{error} field) is $0.0\%$ in all nine cells.

\begin{table}[H]
\centering
\small
\caption{Per-arm overhead for the three-arm mechanism ablation across three datasets. The ctx-inject arm performs the same trigger detection as \rg but appends the rank-1 chunk text as a \texttt{user}-role observation instead of forcing a \texttt{read} tool call.}
\label{tab:r6-overhead}
\setlength{\tabcolsep}{3pt}
\resizebox{\linewidth}{!}{%
\begin{tabular}{llrrrrr}
\toprule
Dataset & Arm & LLM-Acc & Reads/Q & CtxInj/Q & Loops/Q & Cost/Q \\
\midrule
\multirow{3}{*}{HotpotQA}
 & (i) no-\rg       & 79.6 & 0.81 & ---  & 3.07 & \$0.0052 \\
 & (ii) \rg         & \textbf{82.8} & 1.03 & ---  & 3.63 & \$0.0020 \\
 & (iii) ctx-inject & 79.5 & 0.85 & 0.41 & 3.47 & \$0.0019 \\
\midrule
\multirow{3}{*}{2Wiki}
 & (i) no-\rg       & 64.4 & 0.77 & ---  & 3.54 & \$0.0082 \\
 & (ii) \rg         & \textbf{69.7} & 1.06 & ---  & 4.32 & \$0.0054 \\
 & (iii) ctx-inject & 57.0 & 0.74 & 0.76 & 4.00 & \$0.0045 \\
\midrule
\multirow{3}{*}{MuSiQue}
 & (i) no-\rg       & 34.2 & 0.35 & ---  & 2.90 & \$0.0057 \\
 & (ii) \rg         & \textbf{43.6} & 1.00 & ---  & 4.97 & \$0.0088 \\
 & (iii) ctx-inject & 38.1 & 0.47 & 0.96 & 3.87 & \$0.0042 \\
\bottomrule
\end{tabular}%
}
\end{table}

\paragraph{Per-arm McNemar tests and effect sizes.}
Across the three paired samples, the \rg{}--vs--no-\rg{} contrast (arm (ii) vs (i)) is highly significant on every dataset: $+3.2$~pp on HotpotQA (paired bootstrap $[+2.0, +4.5]$; exact two-sided McNemar $p=1.9{\times}10^{-7}$), $+5.3$~pp on 2WikiMultiHopQA ($[+3.7, +6.9]$; $p=3.2{\times}10^{-11}$), and $+9.4$~pp on MuSiQue ($[+7.2, +11.6]$; $p=2.3{\times}10^{-21}$). The ctx-inject--vs--no-\rg{} contrast (arm (iii) vs (i)) is $-0.1$~pp on HotpotQA ($[-1.9, +1.7]$; $p=1.0$), $-7.4$~pp on 2Wiki ($[-9.5, -5.4]$; $p=8.4{\times}10^{-9}$), and $+3.9$~pp on MuSiQue ($[+1.7, +6.1]$; $p=0.003$). The direct (iii)~vs~(ii) contrast favors \rg{} on every dataset (HotpotQA $b=20$ ctx-wins, $c=53$ \rg-wins, exact $p=0.000142$; 2Wiki $b=27$, $c=154$, $p=8.5{\times}10^{-23}$; MuSiQue $b=49$, $c=104$, $p=1.0{\times}10^{-5}$). Delivering the same chunk text without the \texttt{read} action therefore recovers at most a minority of \rg's gain (point estimate $+3.9$~pp on MuSiQue vs \rg's $+9.4$~pp), and on 2Wiki it is substantially net-negative (ctx-inject is $7.4$~pp \emph{worse} than no-\rg).

\paragraph{Disagreement composition.}
On the paired HotpotQA sample, arm (iii) and arm (ii) disagree on $73$ trajectories: arm (ii) is correct and arm (iii) is wrong on $53$ ($73\%$), and arm (iii) is correct and arm (ii) is wrong on $20$ ($27\%$). The deflection-pattern rate (final answers containing conversational hedges such as \emph{``thanks''}, \emph{``do you want me to re-evaluate''}, \emph{``please specify''}) is $21/53 = 40\%$ in arm-(iii)-wrong responses on these disagreement cases, and $0/53 = 0\%$ in the corresponding arm-(ii)-correct responses. A representative case is HotpotQA \texttt{5a84e0a45542991dd0999e11} (\emph{``Which is farther north, Steel Venom or Wicked Twister?''}): under \rg, the agent reads chunk 29 and answers \emph{``Wicked Twister''} correctly; under ctx-inject, the agent receives the identical chunk-29 text in its message history and replies \emph{``Thanks --- I've noted that additional context for chunk 29. Do you want me to re-evaluate \ldots?''}, never committing to an answer. A second case (\texttt{5a7c8a3b55429935c91b5204}) shows the opposite direction --- ctx-inject succeeds on a question \rg got wrong --- and contributes to the $20$ ctx-wins. Full case study and analysis script in \texttt{revision\_analysis/r6\_context\_injection.md}.

\subsection{Channel-vs-Action Decomposition (Four-Arm Probe)}
\label{app:toolrole}

The three-arm ablation above delivers ctx-inject text as a \texttt{user}-role observation, leaving open whether \rg's advantage reflects the \texttt{tool}-role delivery channel or the self-issued \texttt{read} action itself. We add a fourth arm on MuSiQue ($n=500$, gpt-5-mini minimal, same paired question IDs, seed 42) that injects the identical rank-1 chunk text as a fabricated \texttt{tool}-role response (a synthetic read tool-call/result pair) without the agent issuing the corresponding action (Table~\ref{tab:toolrole}).

\begin{table}[H]
\centering
\small
\caption{Four-arm channel-vs-action probe (MuSiQue, gpt-5-mini minimal, paired $n=500$, seed 42). $\Delta$ is LLM-Acc relative to no-\rg{} (34.8).}
\label{tab:toolrole}
\begin{tabular}{lrrl}
\toprule
Arm & LLM-Acc & $\Delta$ & 95\% CI \\
\midrule
no-\rg                   & 34.8 & ---             & ---            \\
ctx-inject (user-role)   & 39.2 & $+4.4$          & $[+0.8,+7.8]$  \\
ctx-inject (tool-role)   & 40.0 & $+5.2$          & $[+2.4,+8.8]$  \\
\rowcolor{rowhi}\rg      & \textbf{43.0} & $\mathbf{+8.2}$ & $[+4.6,+11.6]$ \\
\bottomrule
\end{tabular}
\end{table}

Channel alone (tool-role vs.\ user-role injection, neither self-issued) accounts for $+5.2-4.4=+0.8$~pp, with heavily overlapping CIs --- not significant. The self-issued read action (\rg{} vs.\ tool-inject, same channel) adds a further $+3.0$~pp. \rg's advantage is therefore most consistent with the agent committing to a self-issued read action rather than with the \texttt{tool}-role delivery channel --- an action-commitment interpretation that these controls support but do not conclusively isolate --- refining, rather than contradicting, the three-arm result above: forced context in either channel recovers at most about half of \rg's gain, and channel choice is not the active ingredient. This probe uses a single dataset and backbone (MuSiQue was chosen for its large \rg headroom); the three-arm result across all three datasets (Table~\ref{tab:mechanism-ablation}) remains the primary mechanism evidence.

\section{Robustness Details}
\label{app:robustness}

This appendix expands the robustness checks summarized in \S\ref{sec:exp-robustness}.

\paragraph{Threshold and priority.}
Varying the low-evidence entity-coverage threshold from 0.6 to 0.9 barely changes minimal-reasoning discipline rates: HotpotQA ranges from 12.4--13.3\%, 2WikiMultiHopQA from 21.9--22.1\%, and MuSiQue from 56.9--57.0\%. Changing the priority order changes absolute subtype counts, but a strict no-read-only lower bound preserves the same minimal-reasoning hump (HotpotQA 9.3\%, 2WikiMultiHopQA 15.4\%, MuSiQue 51.5\%). Because the fine-grained subtype boundaries depend on the entity-coverage threshold, we treat them as approximate and base the H1 overlap result on the coarser binary discipline indicator, whose cross-extractor agreement we quantify next.

\paragraph{Extractor robustness for the discipline/post-gold-read overlap.}
Table~\ref{tab:extractor-overlap} gives the 2$\times$4 bucket shares over the 3{,}807 wrong cases under regex and spaCy \texttt{en\_core\_web\_sm} entity extractors. The two extractors agree on the binary discipline indicator at Cohen's $\kappa=0.628$ (observed agreement $82.1\%$; expected $52.0\%$). The H1 conclusion is not the precise value of any single share but the qualitative two-axis structure, and that structure is invariant to extractor choice: under both extractors the both-trigger overlap stays in a tight $[11.2\%, 13.1\%]$ band---far below the $60\%$ level at which the two axes would collapse into one---and the discipline-only share exceeds the post-gold-read-only share by more than two to one. The residual extractor disagreement is concentrated in borderline coverage cases near the $0.8$ threshold; because spaCy and regex shift such cases in the \emph{same} direction (spaCy raises both the discipline-only and both-trigger shares), the rank order and the sub-threshold overlap are preserved rather than reshuffled.

\begin{table}[H]
\centering
\small
\caption{Multi-label buckets over the 3{,}807 wrong cases, by entity extractor (regex / spaCy \texttt{en\_core\_web\_sm}). Cohen's $\kappa=0.628$.}
\label{tab:extractor-overlap}
\setlength{\tabcolsep}{4pt}
\begin{tabular}{lrrrr}
\toprule
Extractor & disc-only & post-only & both & neither \\
\midrule
regex   & 46.5\% & 21.4\% & 11.2\% & 20.9\% \\
spaCy   & 50.2\% & 19.5\% & 13.1\% & 17.2\% \\
\bottomrule
\end{tabular}
\end{table}

\paragraph{Bootstrap intervals.}
We compute paired bootstrap 95\% CIs over question IDs (1{,}000 iterations, seed 42). The minimal-reasoning \rg accuracy gains exclude zero on all three datasets: HotpotQA $+3.2$ pp $[+2.0,+4.5]$, 2WikiMultiHopQA $+5.3$ pp $[+3.8,+6.9]$, MuSiQue $+9.4$ pp $[+7.5,+11.5]$. Bootstrap estimates for the induced-read contrast on the zero-read subset are consistent with the point estimates in Table~\ref{tab:induced-read}: HotpotQA $+14.7$ pp $[+9.2,+20.2]$, 2WikiMultiHopQA $+20.1$ pp $[+14.4,+25.4]$, MuSiQue $+14.4$ pp $[+11.5,+17.3]$. The Gemini thinking-budget paired effect is $-4.4$ pp $[-7.3,-1.3]$.

\paragraph{Stratified label permutation.}
The label-permutation test reported in \S\ref{sec:exp-differential} uses a within-cell-stratified null draw over all 12{,}000 trajectories; this preserves cell-level marginals while randomizing the discipline / post-gold-read label assignment. The observed statistic exceeds every one of the 10{,}000 permutation samples, giving empirical $p<10^{-4}$ that is robust to cell-marginal differences.

\paragraph{Multiple-testing adjustment.}
\label{app:fdr}
We apply Benjamini--Hochberg (BH) FDR and Bonferroni corrections across the 11 tests supporting Table~\ref{tab:differential-response}: three \rg{} main effects (\pdisc, \ppost{} marginal, correctness), six effort contrasts (min and med vs.\ gpt-4o-mini, on \pdisc{}/\ppost{}/correctness), the effort$\times$axis interaction reported in Table~\ref{tab:differential-response}, and the label-permutation exchangeability test. Nine GLM-based tests have original $p<10^{-10}$ (BH $q$ and Bonferroni $p$ both $<10^{-10}$); effort (min vs.\ 4o)$\to\ppost{}$ has $p<10^{-4}$ (adjusted similarly); the permutation test reaches its 10{,}000-sample resolution limit ($p<10^{-4}$; Bonferroni $p=0.0011$). All 11 survive both corrections at $\alpha=0.05$. Surviving FDR is necessary but not sufficient: the marginal \rg{}$\to\ppost{}$ OR is significant after correction yet mechanically exposure-confounded, so the within-strata row in Table~\ref{tab:differential-response} is the appropriate behavioral estimate (R-2).

\subsection{Cross-Family Generalization (Qwen2.5)}
\label{app:qwen-generalization}

The main H2 result (\S\ref{sec:exp-conditional}) is established on the OpenAI controller family. As a generalization check, we run the same two-tool interface, retrieval index, top-$k$, and loop budget with open-weight Qwen2.5-Instruct controllers \citep{qwen2.5} (3B and 7B on all three datasets, plus 14B on HotpotQA; gpt-5-mini judge; matched $n=200$ per cell, \rg on/off). Table~\ref{tab:qwen-generalization} reports $\edisc$ and the paired \rg{} $\Delta$ for each cell.

\begin{table}[H]
\centering
\small
\caption{Qwen2.5 cross-family cells (matched $n=200$, gpt-5-mini judge). $\edisc$ is the no-\rg{} discipline-failure rate; $\Delta$\,\rg{} is the paired LLM-Acc change. Only musique-3B is individually significant at this sample size.}
\label{tab:qwen-generalization}
\setlength{\tabcolsep}{4pt}
\begin{tabular}{llrrlr}
\toprule
Model & Dataset & $\edisc$ (\%) & $\Delta$\,\rg & 95\% CI & McNemar $p$ \\
\midrule
3B  & HotpotQA & 51.0 & $-2.0$ & $[-8.0,+4.0]$  & 0.627 \\
7B  & HotpotQA & 10.5 & $-1.0$ & $[-4.5,+2.5]$  & 0.774 \\
14B & HotpotQA & 31.5 & $-2.5$ & $[-8.0,+2.5]$  & 0.442 \\
\midrule
3B  & 2Wiki    & 78.0 & $+1.5$ & $[-4.0,+7.5]$  & 0.743 \\
7B  & 2Wiki    & 17.5 & $-2.0$ & $[-8.0,+3.5]$  & 0.597 \\
\midrule
3B  & MuSiQue  & \textbf{74.0} & \cellcolor{cellhi}$\mathbf{+6.0}$ & $[+1.0,+11.5]$ & \textbf{0.043} \\
7B  & MuSiQue  & 19.5 & $+1.0$ & $[-4.0,+6.0]$  & 0.845 \\
\bottomrule
\end{tabular}
\end{table}

Across the 7 cells, $\edisc$ and $\Delta$\,\rg{} correlate at Pearson $r=+0.629$ (Spearman $\rho=0.393$), close to the OpenAI-family value of $+0.728$ reported in \S\ref{sec:exp-conditional}: high-$\edisc$ cells (musique-3B, 2wiki-3B) get the largest \rg{} gains, and low-$\edisc$ cells cluster near zero or negative, reproducing the same qualitative dose-response shape on an independent model family.

This is a direction, not a power, result. At $n=200$/cell most individual cells are not independently significant; only musique-3B is ($+6.0$~pp $[+1.0,+11.5]$, $p=0.043$), and all three HotpotQA cells are flat or slightly negative regardless of model size. The claim this table supports is narrow: the \emph{framework's structural relationship} --- \rg's gain scales with residual discipline error --- generalizes across model families. It is not a claim that \rg{} helps uniformly; on this family, as on the OpenAI family's medium-reasoning cells, low-$\edisc$ operating points show no benefit or a small cost.

\section{Gemini Representative Trace and Break Decomposition}
\label{app:gemini-example}

\paragraph{Break decomposition.} The break decomposition is consistent across both datasets: roughly half of the breaks are direct thinking-caused zero-read transitions (HotpotQA $41.9\%$, 2Wiki $50.5\%$), and roughly half retain the same read count but produce a different wrong answer (HotpotQA $50.0\%$, 2Wiki $45.5\%$). The 2Wiki effect is in fact stronger than on HotpotQA, with a four-fold larger zero-read shift, suggesting the failure mode is robust rather than dataset-specific within Wikipedia-style multi-hop QA. Headline zero-read shifts: HotpotQA $52.3\% \to 58.0\%$ ($+5.7$~pp); 2WikiMultiHopQA $40.8\% \to 65.6\%$ ($+24.8$~pp).

\paragraph{Truncation artifact.} A subset of the same-read breaks on 2Wiki are mid-sentence truncations driven by a \texttt{thinking\_budget} $\times$ \texttt{max\_output\_tokens} interaction at the API surface, rather than evidence-skipping; the headline zero-read transition rate ($50.5\%$ of breaks) is unaffected by this artifact.

\paragraph{Representative trace.} A representative Gemini break is the question \textit{``What is the genus of the viral disease that has symptoms such as fever, chills, loss of appetite, nausea, muscle pains, and headaches, and has a chance of causing liver damage?''} (gold answer: \textit{Flavivirus}). With thinking disabled, the agent searches, expands a chunk that links yellow-fever symptoms to the genus, and answers \textit{Flavivirus} correctly (one read, three loops). With a 1024-token thinking budget, the agent emits 1024 internal thinking tokens, searches once, and finalizes without any \texttt{read} action; the answer text restates the symptoms but never names the genus. The trajectory is wrong despite spending more compute.

\section{Procedural Gate-Family Probe}
\label{app:gate-family}
\rg is the minimal procedural gate used in the main experiments. We also ran a small appendix probe on broader deterministic gates to measure the trade-off from adding stronger runtime invariants (Table~\ref{tab:gate-family-probe}). The variants reuse the same saved detector checks: \texttt{no\_read} blocks final answers before any read; \texttt{snippet} additionally blocks answers whose entities appear only in search snippets; \texttt{lowev} additionally blocks final answers when read text has low question-entity coverage; \texttt{full} adds a format-mismatch check. These variants use rules-only corrective hints and no learned controller.

\begin{figure*}[!t]
    \centering
    \maybegraphic[width=0.85\textwidth]{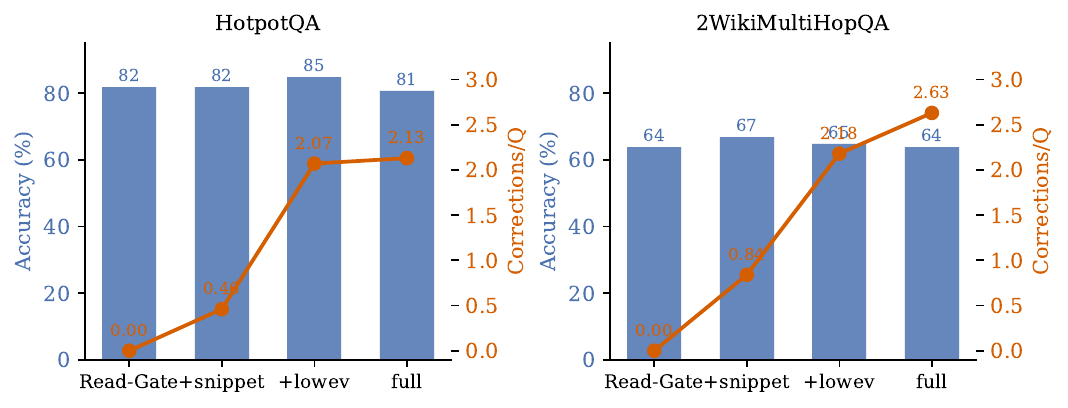}
    \caption{Precision--intervention trade-off across gate variants. Bars: LLM-Acc; lines: Corrections/Q. \texttt{+snippet} stays below 1 Corr./Q with neutral-to-positive accuracy; \texttt{+lowev} and \texttt{full} exceed 2 Corr./Q for accuracy within $\pm 4$ pts of \rg. Matched $n=100$.}
    \label{fig:gate-family-tradeoff}
\end{figure*}

\begin{table*}[!t]
\centering
\small
\caption{Gate-family probe on matched $n=100$ samples (smaller than main-cell tables \ref{tab:rg-overhead}--\ref{tab:contain-acc}). Corr./Q is mean interventions per question; Net vs.\ \rg{} is paired fixes minus breaks. Broader gates help on some cells but are intervention-heavy and dataset-sensitive.}
\label{tab:gate-family-probe}
\begin{tabular}{llrrrr}
\toprule
Dataset & Variant & Acc & Zero-read & Corr./Q & Net vs.\ \rg \\
\midrule
HotpotQA & \rg ($n=100$)      & 82.0 & 0.0 & 0.00 & --   \\
HotpotQA & no-read + snippet  & 82.0 & 0.0 & 0.46 & $0$  \\
HotpotQA & no-read + lowev    & 85.0 & 0.0 & 2.07 & $+3$ \\
HotpotQA & full               & 81.0 & 0.0 & 2.13 & $-1$ \\
\midrule
2Wiki    & \rg ($n=100$)      & 64.0 & 0.0 & 0.00 & --   \\
2Wiki    & no-read + snippet  & 67.0 & 0.0 & 0.84 & $+3$ \\
2Wiki    & no-read + lowev    & 65.0 & 0.0 & 2.18 & $+1$ \\
2Wiki    & full               & 64.0 & 0.0 & 2.63 & $0$  \\
\bottomrule
\end{tabular}
\end{table*}

The result supports the conservative main design choice. As Figure~\ref{fig:gate-family-tradeoff} makes visible, a snippet-only extension is relatively stable, but low-evidence and format gates introduce a stronger precision--intervention trade-off. We therefore use \rg as the main intervention: it is the weakest runtime invariant that directly targets pre-evidence termination without adding heuristic coverage thresholds or a learned controller.

\section{Entity-Coverage Labeling}
\label{app:entity-labeling}

The discipline subtypes \texttt{snippet-only final} and \texttt{low-evidence final} require detecting whether question-side named entities appear in retrieved snippets and read chunks (\S\ref{sec:failure-indicators}). We use a deterministic matching pipeline. When spaCy is available, we run the \texttt{en\_core\_web\_sm} pipeline on the question, retrieved snippets, and read chunks, and treat its named-entity spans as coverage candidates. When spaCy is unavailable or returns no spans, we fall back to a regex that captures capitalized multi-word phrases (\texttt{[A-Z][A-Za-z0-9.\char`\-]+(?:\textbackslash s[A-Z][A-Za-z0-9.\char`\-]+)*}) and standalone numeric tokens, which together cover proper-noun answer entities and date or count answers common in HotpotQA / 2WikiMultiHopQA / MuSiQue. Matching is case-insensitive over the union of all question entities; an entity is covered if any of its surface forms appears verbatim in the candidate text. Coverage rate is the fraction of question entities matched in read chunks; a trajectory is labeled \texttt{low-evidence final} when this rate falls below 0.8. Because subtype assignment depends on this coverage threshold, the fine-grained \texttt{snippet-only}/\texttt{low-evidence} split is approximate; the H1 overlap analysis in Appendix~\ref{app:robustness} therefore rests on the coarser binary discipline indicator, on which the two extractor paths agree at Cohen's $\kappa=0.628$.

\subsection{Entity-Matcher Validation}
\label{app:entity-matcher-validation}

We validate the regex entity matcher against an author-verified, hand-labeled sample ($n=60$: 30 questions and 30 gold reference answers, stratified across HotpotQA and 2WikiMultiHopQA; every label reviewed by an author). Table~\ref{tab:entity-matcher} reports precision, recall, and F1 under contains-match, the same matching rule the discipline detector uses operationally. Precision is $1.000$ (no false positives among 69 predicted entities; 95\% Wilson CI $[0.947,1.000]$); the recall gap ($0.734$ overall) is concentrated almost entirely in single-token entities (recall $0.148$), which the regex cannot match by construction (it requires $\geq 2$ consecutive capitalized tokens), versus $0.966$ recall on the multi-word entities it targets. Because the matcher produced no spurious entity on this sample, the reported discipline-failure rates are a conservative lower bound rather than an inflated estimate.

\begin{table}[H]
\centering
\small
\caption{Entity-matcher validation against an author-verified hand-labeled sample ($n=60$), contains-match.}
\label{tab:entity-matcher}
\begin{tabular}{lrrr}
\toprule
 & Precision & Recall & F1 \\
\midrule
Overall                & 1.000 & 0.734 & 0.847 \\
Single-token entities  & ---   & 0.148 & ---   \\
Multi-word entities    & ---   & 0.966 & ---   \\
\bottomrule
\end{tabular}
\end{table}

\section{Discipline-Failure Subtype Decomposition}
\label{app:subtype-decomposition}

Section~\ref{sec:failure-indicators} defines three discipline-failure subtypes: no-read final, snippet-only final, and low-evidence final. Table~\ref{tab:subtype-decomposition} reports their relative share, as a percentage of $\edisc$ (the discipline failures in that cell), across the nine main dataset$\times$backbone cells.

\begin{table}[H]
\centering
\small
\caption{Discipline-failure subtype shares (\% of $\edisc$ in that cell) across the nine main cells.}
\label{tab:subtype-decomposition}
\setlength{\tabcolsep}{4pt}
\begin{tabular}{llrrrr}
\toprule
Dataset & Backbone & $n_{\edisc}$ & No-read & Snippet-only & Low-ev. \\
\midrule
HotpotQA & 4o-mini   & 76  & 3.9  & 9.2  & 86.8 \\
HotpotQA & 5mini-min & 133 & 69.9 & 1.5  & 28.6 \\
HotpotQA & 5mini-med & 45  & 37.8 & 6.7  & 55.6 \\
2Wiki    & 4o-mini   & 128 & 7.8  & 11.7 & 80.5 \\
2Wiki    & 5mini-min & 221 & 69.7 & 3.6  & 26.7 \\
2Wiki    & 5mini-med & 57  & 38.6 & 26.3 & 35.1 \\
MuSiQue  & 4o-mini   & 278 & 15.5 & 17.6 & 66.9 \\
MuSiQue  & 5mini-min & 570 & 90.4 & 2.1  & 7.5  \\
MuSiQue  & 5mini-med & 217 & 56.7 & 22.6 & 20.7 \\
\midrule
\multicolumn{3}{l}{\textbf{Pooled (1{,}725 discipline failures)}} & \textbf{56.8} & \textbf{9.3} & \textbf{33.9} \\
\bottomrule
\end{tabular}
\end{table}

Pooled over all nine cells, $56.8\%$ of discipline failures are strict no-read finals, $9.3\%$ take the prompt-permitted snippet-only path, and $33.9\%$ are low-evidence. Because it requires no entity-matching heuristic, no-read is the primary, non-heuristic discipline signal; the two NER-based subtypes are secondary diagnostics whose relative weight is backbone-dependent (under gpt-4o-mini, which rarely skips reads outright, low-evidence dominates instead). Notably, $43.2\%$ of discipline failures ($745/1{,}725$) occur despite $\text{read\_count}\geq 1$ (snippet-only plus low-evidence): \rg is framed as removing the zero-read floor, not as guaranteeing adequate post-read inspection.

\section{Zero-Read Provenance: Leakage vs.\ Unsupported Guesses}
\label{app:zero-read-leakage}

Using the exact chunk IDs each \texttt{search} call returned (already logged per trajectory), we check whether a zero-read answer's predicted entity is visible anywhere in the retrieved-but-unread search results, separating snippet leakage from genuinely unsupported guesses.

\paragraph{Wrong zero-read answers.} Pooled across the nine non-\rg main cells ($n=980$ wrong zero-read trajectories, 300-character snippet definition): $23.7\%$ (232) show snippet leakage (the predicted entity appears in a retrieved-but-unread chunk), $58.5\%$ (573) show no visible evidence (no trace of the guessed entity in anything retrieved), $5.1\%$ (50) never called \texttt{search} at all, and the remaining $12.8\%$ (125) yield no informative entity in the predicted answer to match against (e.g., yes/no or clarification-style answers). The four categories are mutually exclusive and sum to $100\%$. Leakage is a real but minority contributor; most zero-read errors are unsupported guesses rather than surfaced-but-ignored answers.

\paragraph{Correct zero-read answers.} Restricted to the HotpotQA+2WikiMultiHopQA gold-annotated subset ($n=282$ classifiable correct zero-read cases), $19.5\%$ (55) show clean snippet leakage of the correct entity, $68.4\%$ (193) retrieved the gold-supporting chunk but without a verbatim entity match (a candidate for genuine cross-snippet synthesis without a full read), and $12.1\%$ (34) show no retrieved connection to gold evidence at all, consistent with parametric knowledge. Leakage is again a minority explanation, on both the wrong and correct sides.

Critically, leakage is not the same question as whether a read action occurred: even when the answer surfaces in a snippet, finalizing without reading still skips the verification step our framework measures.

\paragraph{Snippet-length sensitivity.} Because ``snippet'' is an operational choice, Table~\ref{tab:leakage-sensitivity} sweeps the char-prefix definition from 100 characters to the full chunk. Leakage rises with a more generous definition, from $12.2\%$ to $60.3\%$ on wrong zero-read answers; we report the conservative 300-character definition ($23.7\%$) as the headline, with this sweep reported as a bounded robustness range rather than a point estimate.

\begin{table}[H]
\centering
\small
\caption{Snippet-length sensitivity sweep (char-prefix definition; 9 non-\rg main cells).}
\label{tab:leakage-sensitivity}
\begin{tabular}{lrr}
\toprule
Snippet definition & Wrong-ZR leak \% & Correct-ZR leak \% \\
\midrule
100 chars/chunk           & 12.2 & 8.7  \\
300 chars/chunk (default) & 23.7 & 21.1 \\
600 chars/chunk           & 30.9 & 32.6 \\
1200 chars/chunk          & 40.8 & 44.4 \\
Full chunk                & 60.3 & 73.0 \\
\bottomrule
\end{tabular}
\end{table}

\section{Differential Intervention Visualization}
\label{app:intervention-plane}

\begin{figure*}[!t]
    \centering
    \maybegraphic[width=0.85\textwidth]{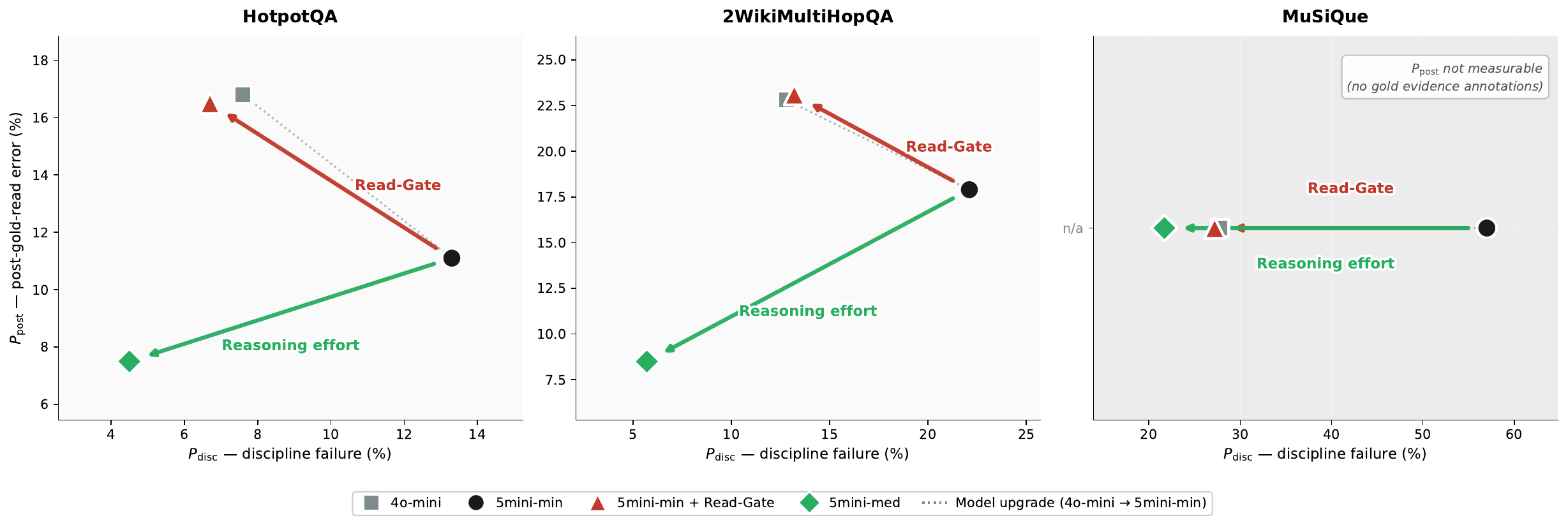}
    \caption{Differential intervention signatures in the $(\pdisc,\ppost)$ plane. \rg{} mainly reduces pre-evidence failures; higher effort moves both indicators with a different slope. Markers: \textbf{4o}=gpt-4o-mini; \textbf{min}/\textbf{med}=gpt-5-mini minimal/medium; \textbf{min+\rg}=min with \rg. MuSiQue's $\ppost\!=\!0$ (no gold).}
    \label{fig:intervention-plane}
\end{figure*}

Figure~\ref{fig:intervention-plane} visualizes the intervention signatures in the $(\pdisc,\ppost)$ plane. The corresponding odds ratios and tests are reported in Table~\ref{tab:differential-response} in the main text.

\section{Prompt-Only Control}
\label{app:prompt-control}

Table~\ref{tab:prompt-control} reports the prompt-only baseline summarized in
\S\ref{sec:exp-readgate}, extended to all three datasets (gpt-5-mini minimal,
paired by question ID). The strict-prompt variant removes the snippet
exception from the system prompt and explicitly forbids answering before a
successful read, but uses no environment-level gate. Across the three
datasets, the strict prompt reduces zero-read trajectories, but its accuracy
effect is unreliable: $+0.0$~pp on HotpotQA, $-2.8$~pp on 2WikiMultiHopQA,
and $+3.2$~pp on MuSiQue. \rg, by contrast, recovers $+3.2$/$+5.3$/$+9.4$~pp
on the same cells and reduces zero-read to $\leq\!3.8\%$ everywhere. On every
dataset, the paired (RG vs strict-prompt) contrast is significant
(McNemar $p \leq 0.005$).

\begin{table}[H]
\centering
\small
\caption{Prompt-only control across three datasets, gpt-5-mini minimal. Prompting reduces zero-read behavior on all datasets but does not match \rg's accuracy gain on any dataset, and is net-negative on 2Wiki. McNemar $p$ is the paired \rg vs strict-prompt LLM-Acc test.}
\label{tab:prompt-control}
\setlength{\tabcolsep}{3.5pt}
\resizebox{\linewidth}{!}{%
\begin{tabular}{llrrl}
\toprule
Dataset & Condition          & Zero-read & LLM-Acc & vs strict $p$ \\
\midrule
\multirow{3}{*}{HotpotQA}
 & Original prompt          & 21.8\%    & 79.6   & --- \\
 & Strict prompt only       & 12.4\%    & 79.6   & --- \\
 & \cellcolor{rowhi}\textbf{\rg} & \cellcolor{rowhi}\textbf{0.3\%} & \cellcolor{rowhi}\textbf{82.8} & \cellcolor{rowhi}$0.0048$ \\
\midrule
\multirow{3}{*}{2Wiki}
 & Original prompt          & 26.4\%    & 64.4   & --- \\
 & Strict prompt only       & 20.7\%    & 61.6   & --- \\
 & \cellcolor{rowhi}\textbf{\rg} & \cellcolor{rowhi}\textbf{0.3\%} & \cellcolor{rowhi}\textbf{69.7} & \cellcolor{rowhi}$2.5{\times}10^{-7}$ \\
\midrule
\multirow{3}{*}{MuSiQue}
 & Original prompt          & 65.5\%    & 34.2   & --- \\
 & Strict prompt only       & 48.1\%    & 37.4   & --- \\
 & \cellcolor{rowhi}\textbf{\rg} & \cellcolor{rowhi}\textbf{3.8\%} & \cellcolor{rowhi}\textbf{43.6} & \cellcolor{rowhi}$1.5{\times}10^{-5}$ \\
\bottomrule
\end{tabular}}
\end{table}

\section{Cross-Judge Robustness Details}
\label{app:cross-judge}

This appendix expands the cross-judge robustness paragraph in \S\ref{sec:evaluation-protocol}. We re-judged $n=450$ stored (question, gold, prediction) triples (75 per cell across HotpotQA, 2WikiMultiHopQA, and MuSiQue $\times$ no-\rg / \rg) under Gemini 2.5 Pro (temperature 0.0, $\mathrm{thinking\_budget}=256$), using the verbatim judge prompt in Appendix~\ref{app:judge-prompt}. The sample is stratified by the gpt-5-mini-judge agreement landscape between no-\rg and \rg on the same question (50\% disagreement, 30\% both-correct, 20\% both-incorrect) so that the $\kappa$ statistic is informed by the cases most likely to flip across judges. Stratum-wise raw agreement is $0.952$ on the disagreement stratum (228 judgments), $0.970$ on both-correct (132), and $0.978$ on both-incorrect (90).

\paragraph{Per-cell $\kappa$ and reweighted LLM-Acc.}
Table~\ref{tab:cross-judge} reports Cohen's $\kappa$ alongside the stratum-reweighted Gemini LLM-Acc and its gap to the paper headline. All six cells fall in the ``almost perfect'' band ($\kappa\geq 0.80$); the reweighted Gemini estimates are within $\pm 3.7$ pp of the paper headline on every cell. The largest gaps are uniform shifts across conditions that cancel in $\Delta$ (Gemini HotpotQA $\Delta = +3.2$, identical to the paper; Gemini 2Wiki $\Delta = +7.4$ vs the paper's $+5.3$; Gemini MuSiQue $\Delta = +7.3$ vs the paper's $+9.4$). The family-bias hypothesis would predict $\Delta_\text{Gemini} \ll \Delta_\text{gpt5mini}$; we observe direction-preserving and similar-magnitude $\Delta$ across all three datasets.

\begin{table}[!t]
\centering
\small
\caption{Cross-judge agreement and stratum-reweighted LLM-Acc on the $n=450$ sample. ``Paper'' is gpt-5-mini on the full $n=1000$ cell; ``Gemini-rw'' is the Horvitz--Thompson-reweighted Gemini 2.5 Pro estimate on the 75-per-cell sample.}
\label{tab:cross-judge}
\setlength{\tabcolsep}{4pt}
\resizebox{\linewidth}{!}{%
\begin{tabular}{lrrrr}
\toprule
Cell & $\kappa$ & Paper & Gemini-rw & Gap \\
\midrule
HotpotQA, no-\rg & 0.940 & 79.6 & 75.9 & $-3.7$ \\
HotpotQA, \rg    & 0.932 & 82.8 & 79.1 & $-3.7$ \\
2Wiki, no-\rg    & 0.886 & 64.4 & 63.6 & $-0.8$ \\
2Wiki, \rg       & 0.871 & 69.7 & 71.0 & $+1.3$ \\
MuSiQue, no-\rg  & 0.970 & 34.2 & 34.2 & $0.0$ \\
MuSiQue, \rg     & 0.860 & 43.6 & 41.5 & $-2.1$ \\
\midrule
Pooled (all 6) & \textbf{0.924} & --- & --- & --- \\
$\Delta$ HotpotQA & --- & $+3.2$ & $+3.2$ & $0.0$ \\
$\Delta$ 2Wiki    & --- & $+5.3$ & $+7.4$ & $+2.1$ \\
$\Delta$ MuSiQue  & --- & $+9.4$ & $+7.3$ & $-2.1$ \\
\bottomrule
\end{tabular}%
}
\end{table}

\paragraph{Disagreement case study.}
gpt-5-mini and Gemini 2.5 Pro disagree on 17/450 cases (3.8\%). The pattern is bidirectional: a small share where Gemini catches a subject-mismatch, internal contradiction, or logic inversion that gpt-5-mini missed; a similar share where gpt-5-mini correctly accepts a paraphrase or partial answer that Gemini penalizes on format strictness; the rest are genuinely borderline.

A representative Gemini-catches case is the 2Wiki question \emph{``Where was the place of death of Bess Taffel's husband?''}, where the prediction confuses subject (gives Bess Taffel's place of death, not her husband's) but contains the token-matching ``Los Angeles''; gpt-5-mini marks it correct, Gemini marks it incorrect. Such cases reflect a slight strictness difference, not a family-bias artifact: the disagreements are small in number, bidirectional, and do not concentrate on gpt-5-mini-family generations. Full case study and reweighting derivation in \texttt{revision\_analysis/r1b\_cross\_judge.md}.

\section{Architecture $\times$ Backbone Comparison}
\label{app:architecture}

The runtime findings in the main text fix the agent interface. We now vary it, comparing the two-tool design against A-RAG's three-tool hierarchical interface (keyword search, semantic search, read) under the same index, top-$k$, loop budget, judge, and reasoning effort (Table~\ref{tab:architecture-comparison}). A-RAG numbers at gpt-5-mini medium are taken from \citet{du2026arag}; A-RAG at gpt-4o-mini and at gpt-5-mini minimal are our re-measurements under the same retrieval index, top-$k$, loop budget, judge, and protocol.

\begin{table}[!t]
\centering
\small
\caption{\textbf{Backbone-conditional interface effect.} Two-tool interface vs.\ A-RAG's three-tool hierarchical interface; $\Delta$ is 2-tool minus A-RAG in LLM-Acc points; $^{*}$ marks paired McNemar $p<0.05$. Same retrieval index, top-$k$, loop budget, and judge throughout. $^{\dagger}$~marks cells cited from \citet{du2026arag}; all other A-RAG cells (4o-mini, 5mini-min rows) are re-measured under our harness.}
\label{tab:architecture-comparison}
\setlength{\tabcolsep}{4pt}
\begin{tabular}{llrrr}
\toprule
Backbone & Dataset & A-RAG & 2-tool & $\Delta$ \\
\midrule
4o-mini   & HotpotQA & 77.6 & 81.6 & \cellcolor{cellpos}$\mathbf{+4.0^{*}}$  \\
4o-mini   & 2Wiki    & 53.8 & 70.1 & \cellcolor{cellpos}$\mathbf{+16.3^{*}}$ \\
4o-mini   & MuSiQue  & 41.4 & 46.6 & \cellcolor{cellpos}$\mathbf{+5.2^{*}}$  \\
\midrule
5mini-min & HotpotQA & 73.6 & 79.6 & \cellcolor{cellhi}$\mathbf{+6.0^{*}}$   \\
5mini-min & 2Wiki    & 67.5 & 64.4 & \cellcolor{cellneg}$-3.1$               \\
5mini-min & MuSiQue  & 40.7 & 34.2 & \cellcolor{cellneg}$-6.5^{*}$           \\
\midrule
5mini-med & HotpotQA & 94.5$^{\dagger}$ & 90.8 & \cellcolor{cellneg}$-3.7$               \\
5mini-med & 2Wiki    & 89.7$^{\dagger}$ & 88.8 & \cellcolor{cellneg}$-0.9$               \\
5mini-med & MuSiQue  & 74.1$^{\dagger}$ & 67.1 & \cellcolor{cellneg}$-7.0^{*}$           \\
\bottomrule
\end{tabular}
\end{table}

The 2-tool 5mini-med row reports the full $n=1{,}000$ no-\rg{} cell mean for this backbone. \S\ref{sec:exp-conditional} additionally reports a matched $n=100$ subsample of the same system, paired question-by-question with the \rg{} medium ablation, giving 95.0/90.0/74.0 on HotpotQA/2Wiki/MuSiQue (Table~\ref{tab:conditional-rg}) --- a consistent measurement of the same no-\rg{} system on a different, smaller question subsample used for the paired \rg{} comparison, not a different system or a labeling error.

The ordering is monotone in backbone strength. On gpt-4o-mini, the two-tool interface wins on all three datasets ($+4.0$, $+16.3$, $+5.2$, all paired-significant). On gpt-5-mini at minimal reasoning, the advantage holds on HotpotQA but reverses on 2WikiMultiHopQA and MuSiQue. On gpt-5-mini at medium reasoning, the hierarchical interface wins on all three datasets ($-3.7$, $-0.9$, $-7.0$). Neither architecture is universally better.

A natural reading is that interface granularity trades decision burden against controllability: the two-tool interface folds the lexical-vs.-semantic choice into the environment via reciprocal-rank fusion, removing one decision axis from the model. A weaker controller benefits from the reduced burden; a stronger controller with more reasoning effort can exploit the extra control axis when it can reliably route each subquery to a retrieval mode. This is complementary to the \rg result: both findings say the right agentic-RAG design is regime-dependent --- simplified interface plus runtime gate at the low-cost operating point, hierarchical interface at the high-reasoning operating point --- without changes to model weights or the retriever.

\end{document}